\documentclass{article}

 \usepackage[eandd, final]{neurips_2026}
\usepackage[utf8]{inputenc} % allow utf-8 input
\usepackage[T1]{fontenc}    % use 8-bit T1 fonts
\usepackage{hyperref}       % hyperlinks
\usepackage{url}            % simple URL typesetting
\usepackage{booktabs}       % professional-quality tables
\usepackage{amsfonts}       % blackboard math symbols
\usepackage{nicefrac}       % compact symbols for 1/2, etc.
\usepackage{microtype}      % microtypography
\usepackage{xcolor}         % colors
\usepackage{wrapfig}
\usepackage{booktabs}
\usepackage{multirow}
\usepackage{pifont}
\usepackage{wrapfig}
\usepackage{xspace}  % 必需，用于\xspace命令
\usepackage{xcolor}
\usepackage{colortbl}
\usepackage{multirow}
\usepackage{multicol}
\usepackage{graphicx}

\def\eg{\emph{e.g}.\xspace}

\newcommand{\gray}[1]{\textcolor{gray}{#1}}
\usepackage[dvipsnames]{xcolor}  % 或
\definecolor{gray}{gray}{0.5}
\usepackage{graphicx}
\usepackage{booktabs}
\usepackage{multirow}
\usepackage{pifont}
\usepackage{amsmath}

\newcommand{\method}{{\fontfamily{ppl}\selectfont
COW}\xspace}

\newcommand{\bench}{{\fontfamily{ppl}\selectfont
CausalSpatial}\xspace}

\newcommand{\Method}{{\fontfamily{ppl}\selectfont
\textbf{\textsc{\underline{C}ausal \underline{O}bject \underline{W}orld Model}}}\xspace}

\title{\bench: A Benchmark for Object-Centric\\Causal Spatial Reasoning}

\author{
  \textbf{Wenxin Ma}$^{1,2,}$\thanks{Equal contribution.} \quad
  \textbf{Chenlong Wang}$^{1,}$\footnotemark[1] \quad
  \textbf{Ruisheng Yuan}$^{1,}$\footnotemark[1] \quad
  \textbf{Hao Chen}$^{1}$ \quad
  \textbf{Nanru Dai}$^{1}$ \\   
  \textbf{Chengxin Qian}$^{1}$ \quad 
  \textbf{Yijun Yang}$^{1}$ \quad
  \textbf{Qi Chen}$^{1}$ \quad
  \textbf{Zhaoyang Wang}$^{1}$ \quad
  \textbf{S. Kevin Zhou}$^{3}$ \\
  \textbf{Jianwen Xie}$^{4}$ \quad
  \textbf{Alan Yuille}$^{1}$ \quad
  \textbf{Jieneng Chen}$^{1,}$\thanks{Corresponding author: \texttt{jienengchen01@gmail.com}} \\[4pt]
  $^{1}$Johns Hopkins University \quad
  $^{2}$UT Austin \quad
  $^{3}$USTC \quad
  $^{4}$Lambda, Inc
}

\begin{document}

\maketitle

\begin{abstract}
Humans can look at a static scene and instantly predict what happens next --- \textit{will moving this object cause a collision?} We call this ability Causal Spatial Reasoning. 
However, current multimodal large language models (MLLMs) cannot do this, as they remain largely restricted to static spatial perception, struggling to answer ``what-if'' questions in a 3D scene.
We introduce \textbf{\bench}, a diagnostic benchmark evaluating whether models can anticipate consequences of object motions across four tasks: Collision, Compatibility, Occlusion, and Trajectory. 
Results expose a severe gap:  humans score 84\% while GPT-5 achieves only 54\%.
Why do MLLMs fail? Our analysis uncovers a fundamental deficiency: models over-rely on textual chain-of-thought reasoning that drifts from visual evidence, producing fluent but spatially ungrounded hallucinations. 
To address this, we propose the \Method (\method), a framework that externalizes the simulation process by generating videos of hypothetical dynamics. 
With explicit visual cues of causality, \method enables models to ground their reasoning in physical reality rather than linguistic priors. 
% Code and dataset are publicly available.
The project page is: \url{https://causalspatial.github.io}.

\end{abstract}

\section{Introduction}
\label{sec:intro}

\begin{figure}[!ht]
    \centering
    \includegraphics[width=\linewidth]{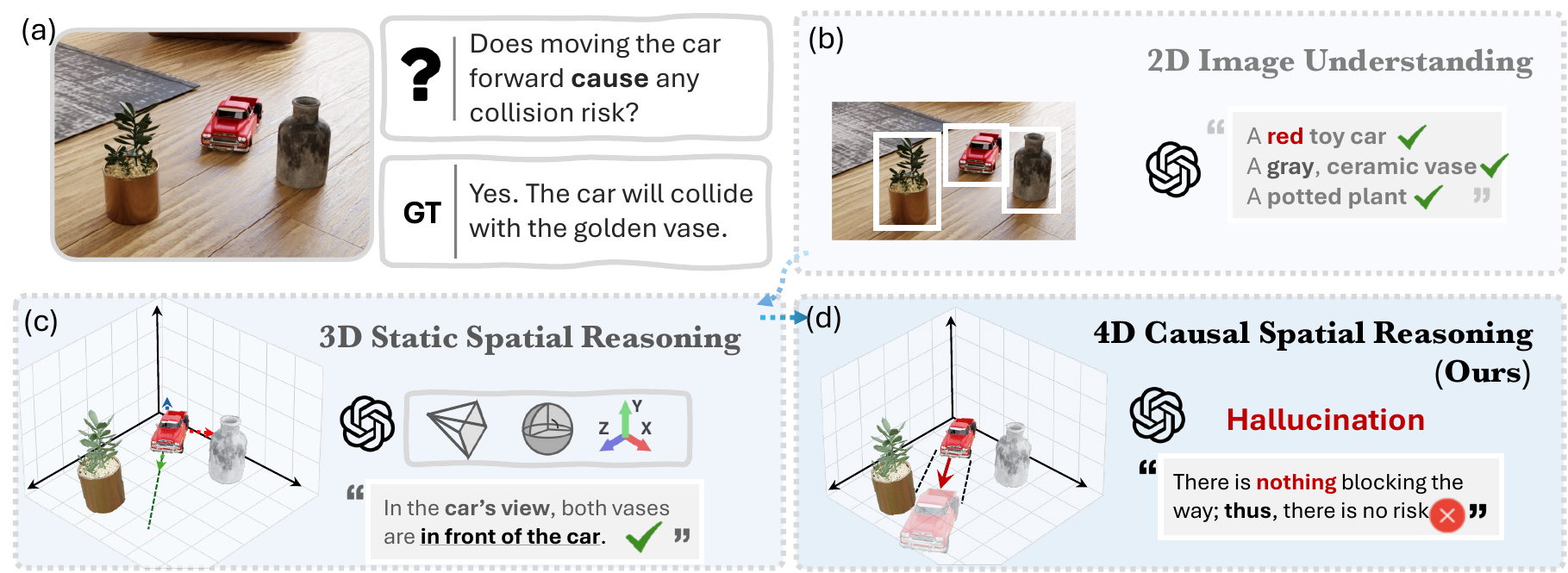}
    \caption{\textbf{Causal spatial reasoning task.} Figure (a) shows an example of the QA pairs. The task requires model to progress from (b) 2D image understanding to (c) 3D spatial perception and ultimately to (d) 4D causal spatial reasoning, predicting how actions affect future spatial configurations. Current MLLMs fail at this complex task due to hallucinations.}
    \label{fig:intro}
\end{figure}

Humans can naturally perceive their surroundings and form a mental 3D model of the world. 
This mental model enables us to simulate possible physical interactions between objects.
For example, as shown in Fig.~\ref{fig:intro}, in a static scene where a car faces a metal vase, we can use spatial cues such as orientation to predict a potential collision if the car moves forward.
We refer to the task of grounding causal inference in spatial reasoning as \textbf{Causal Spatial Reasoning}.

This task requires three progressive spatial reasoning capabilities.
As illustrated in Fig.~\ref{fig:intro}, the process begins with 2D image understanding and 3D static spatial reasoning.
In these paradigms, models are tasked with recognizing objects and estimating spatial cues, such as camera perspective, orientation, and distance.
While these capabilities form the foundation for further reasoning, they are inherently \textit{static}—answering only ``what/where is something'' but failing to address ``what will happen if...'', which is critical for understanding the physical world.
To reason about causality, models must advance to 4D causal spatial reasoning, a comprehensive and dynamic paradigm that integrates spatial understanding with temporal prediction. Rather than analyzing spatial cues in isolation, models must synthesize this information to form coherent mental simulations. This requires first extracting 3D spatial relationships from static images, then simulating hypothetical or counterfactual interactions with temporal consistency (\eg, ``The car will collide with the vase if it moves forward.'').

While humans perform causal spatial reasoning effortlessly, whether MLLMs possess this capability remains largely underexplored.
Current evaluation frameworks (as listed in Tab.\ref{tab:bench-comparison}) are limited to \textit{2D} and \textit{3D} static reasoning or overlook object interactions, failing to assess whether models can reason about how objects dynamically affect each other in the given spatial contexts.

To fill this gap, we propose \bench, a benchmark designed to diagnose causal spatial reasoning capabilities from an object-centric perspective.
Each evaluation item consists of an image showing multiple objects in a scene, paired with a hypothetical object-level motion context (\eg, If the car moves forward, a collision may occur).
This setup requires the model to move beyond static recognition of the given static image, and simulate the future object interaction outcomes according to the motion context.
To provide a comprehensive diagnosis, we define four distinct forms of causal anticipation under different spatial configurations: \textbf{\textit{Collision}}, \textbf{\textit{Occlusion}}, \textbf{\textit{Compatibility}}, and \textbf{\textit{Trajectory}}. 
This design provides a principled decomposition of causal spatial reasoning into core primitives: contact dynamics, visibility changes, geometric constraints, and motion prediction.
As the first open-source diagnostic benchmark of its kind, \bench provides the necessary testbed for advancing models toward more complex and causally grounded spatial understanding.

Extensive experiments on \bench reveal a significant gap between state-of-the-art MLLMs and human performance. 
We carefully attribute this gap to a key limitation: MLLMs tend to rely heavily on textual Chain-of-Thought (CoT) for reasoning about dynamics and causality, which can fail to stay aligned with visual inputs and lead to spatially ungrounded predictions. 
Motivated by this, we hypothesize that explicit visual causality cues can benefit causal spatial reasoning, and explore this hypothesis by proposing \Method (\method). 
\method is an object-centric world model that externalizes causal reasoning by simulating object dynamics as videos. 
This explicit visual evidence helps anchor reasoning in physical reality, enabling more grounded causal inference and improving performance on complex spatial reasoning tasks.
% To address the failure of text-based causal reasoning in maintaining spatial grounding, we propose leveraging world models to provide explicit causality hints with consistent 4D trajectory control.
% We propose \Method (\method), an object-centric world model that externalizes the simulation process by rendering hypothetical object dynamics into video. 
% Unlike textual reasoning, which drifts from reality, \method generates realistic future frames based on extracted 4D trajectories. 
% This explicit visual evidence effectively compensates for the models' inability to perform internal physical simulation, enabling grounded causal inference and demonstrating significant promise in complex spatial reasoning tasks.

In summary, our contribution is three-fold:

\begin{itemize}
\item We introduce \bench, the first object-centric causal spatial reasoning benchmark, comprising diverse tasks. It provides a comprehensive and diagnostic testbed for object-centric causal spatial reasoning.

\item We conduct extensive experiments on \bench across MLLMs, revealing a significant performance gap between machines and humans. 
Our analysis indicates that models hallucinate in textual CoT, struggling to conduct consistent causal reasoning anchored to actual 3D configurations and produce erroneous answers.

\item We hypothesize that explicit visual cues that can help improve grounding in causal spatial reasoning, and propose \method, an object-centric video generation framework that externalizes causality modeling. The results shows the potential of grounding causality in vision.
\end{itemize}
\begin{table}
    \centering
    \newcommand{\yes}{\ding{51}}
    \newcommand{\no}{}
    \caption{\textbf{Comparison of existing benchmarks.}
We compare prior benchmarks across spatial reasoning capabilities and design aspects.
Our benchmark is the only one that jointly supports all evaluation dimensions, enabling comprehensive assessment of object-level causal spatial reasoning.}
    \resizebox{\columnwidth}{!}{%
    \begin{tabular}{l|c|ccccc|cc}
        % \highrow
        \toprule
        \multirow{2}{*}{\textbf{Benchmark}}  & 
        \multirow{2}{*}{\textbf{Domain}} &
        \multicolumn{5}{c|}{\textbf{Evaluation}} &
        \multicolumn{2}{c}{\textbf{Design}}
        \\
        & 
        &
        2D &
        3D &
        Dynamics&
        Imagination &
        Causality &
        Object Ctrl &
        Better Realism \\
        \midrule
        CLEVR-Count~\cite{clevr} &
        Geometric & \no   & \no   & \no   & \no   &    &  &  \\
        CLEVR~\cite{clevr} & 
        Geometric & \yes   & \yes   & \no   & \no  &    &  &  \\
        VQA~\cite{vqa} & 
        Real-world & \yes   & \yes  & \no   & \no  &    &  & \yes \\
        GQA~\cite{hudson2019gqa} &
        Real-world & \yes   & \yes   & \no   & \no  &    &  & \yes \\
        CV-Bench~\cite{cv-bench} &
        Real-world & \yes   & \yes  & \no   & \no  &    &  & \yes \\
        SAT~\cite{ray2024sat} &
        Synthetic & \yes   & \yes   & \yes & \yes &  \yes &  & \no \\
        STI-Bench~\cite{li2025sti} & 
        Real-world & \yes   & \yes   & \yes  &  \no  &  &  & \yes \\
        Spatial457~\cite{spatial457} &
        Vehicles & \yes   & \yes   & \yes  & \yes & \yes & \yes &  \\
        CLEVRER~\cite{yi2019clevrer} &
        Geometric & \yes   & \yes   & \yes   & & \yes      & \yes  &  \\
        CausalVQA~\cite{foss2025causalvqa} &
        Real-world & \yes   & \no   & \yes   &  & \yes    &  & \yes \\
        VSTI-Bench~\cite{vsti-bench} &
        Real-world & \yes   & \yes  & \yes   & &   &  & \yes \\
        WM-Bench~\cite{wm-bench} & 
        Synthetic & \yes   & \yes  & \yes & \yes & \yes &  & \\
        VSI-Super~\cite{yang2025cambrian-s} &
        Real-world  & \yes   & \yes   & \yes & &\no      &  & \yes \\ 
        LEGO-Puzzles~\cite{tang2026lego} & LEGO Bricks & \yes   & \yes & \yes &  \yes & \yes & \yes &\\
        \midrule
        \textbf{\bench} & Synthetic \& Real-world & \yes & \yes & \yes & \yes & \yes & \yes & \yes \\
        \bottomrule
    \end{tabular}
    }
    \label{tab:bench-comparison}
\end{table}

\begin{figure*}[t]
    \centering
    \includegraphics[width=\linewidth]{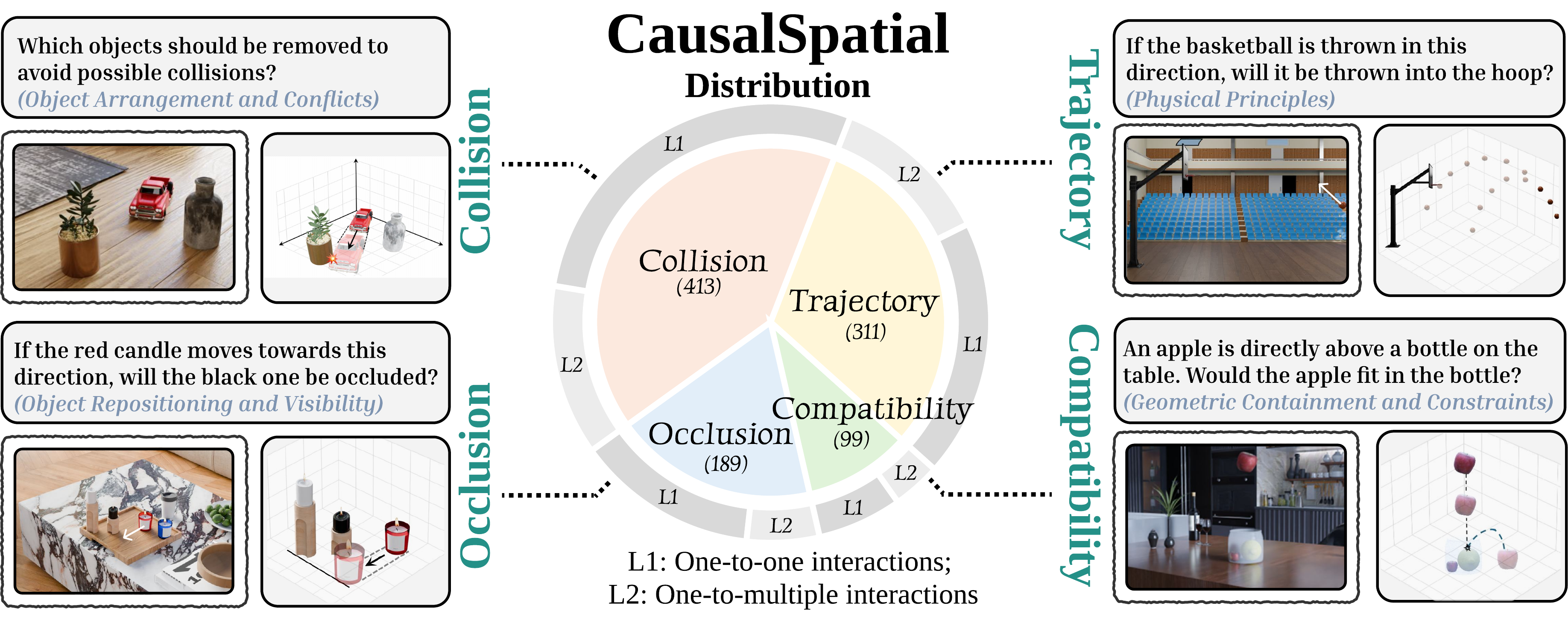}
    % \vspace{-1mm}
    \caption{\textbf{The distribution and tasks of \bench}. It encompasses four causal reasoning tasks that require models to anticipate physical outcomes in 3D scenes: Collision, Occlusion, Trajectory, and Compatibility. The number of evaluation entries for each task is listed in the middle. Each task is designed with two difficulty levels, denoted as ``L1/L2''. All scenes are rendered in Blender to provide realistic 3D environments for physics-grounded evaluation.
    }
    \label{fig:benchmark}
    % \vspace{-3mm}
\end{figure*}

\section{\bench benchmark}
In this section, we formally present \bench, a diagnostic benchmark designed to assess the capability of MLLMs in grounding causal inference within 3D spatial environments. 
We begin by formulating the question of causal spatial reasoning (Sec.~\ref{benchmark:overview}), followed by a detailed taxonomy of the four proposed subtasks (Sec.~\ref{benchmark:taxonomy}). 
We then describe our data collection and annotation pipeline (Sec.~\ref{benchmark:collection}) and conclude with a comprehensive evaluation of current state-of-the-art models (Sec.~\ref{experiment}).

\subsection{Benchmark design}
\label{benchmark:overview}

The design of \bench aims to provide a diagnostic assessment of MLLMs' causal spatial reasoning capabilities. 
We focus on object-centric scenarios in which a specified target object interacts with its surrounding environment under a hypothetical motion. 
Unlike video-based benchmarks that provide temporal observations, \bench requires models to predict future outcomes from a single static scene and a textual intervention, based solely on the underlying 3D spatial relationships. 
This setting \textit{isolates causal simulation from temporal tracking}, reduces shortcut solutions from keyframe cues or temporal correlations~\cite{foss2025causalvqa}, and encourages models to reason about object-grounded spatial interactions rather than relying on superficial pattern matching.

\noindent\textbf{Causality formulation.}
Each query in \bench consists of an image $I_0$ capturing a 3D scene at initial state $S_0$ and a textual motion context $T$ describing a hypothetical causal intervention.
The model is tasked with predicting the resulting future spatial state $S_{\text{future}}$ after this motion is applied.
Unlike traditional VQA, which queries existing static attributes in an image, this task requires counterfactual reasoning about cause-and-effect relationships: how the described motion transforms the initial spatial configuration into a new spatial arrangement, using only the initial visual and textual inputs together with physical world knowledge.

\noindent\textbf{Object-level design.}
A key feature of \bench is its object-level formulation. 
Each motion context focuses on a specific target object and its potential interaction with the surrounding environment. 
This setup decomposes complex dynamics into controlled interaction primitives, requiring the model to: (1) perceive precise 3D spatial attributes of objects in the image (\eg, position, orientation, geometry); (2) perform a mental simulation of the hypothetical motion $m$; and (3) reason about the causal consequences (\eg, collision, occlusion, compatibility, trajectory). 
Such a controlled design enables clearer diagnosis of failure sources, including perception errors, inaccurate dynamics estimation, and ungrounded causal reasoning.

\noindent\textbf{Improved realism.}
Unlike previous benchmarks~\cite{spatial457,clevr,li2023super}, which focus on limited objects categories, simple background, or simple geometric shapes, our benchmark is built on constructed realistic scene settings with diverse object assets and lighting conditions. 
Notably, \bench includes a manually curated real-world subset to evaluate model performance on authentic images captured in uncontrolled environments, providing a complementary validation of real-world causal spatial reasoning. The details of the curation process can be found in Appendix.

\subsection{Tasks}
\label{benchmark:taxonomy}
As shown in Sec.~\ref{fig:benchmark}, we design four types of hypothetical anticipation tasks for systematic evaluation. 
Each task captures a distinct aspect of causal spatial reasoning. 
These motion primitives serve as building blocks of more complex real-world interactions, providing a controlled setting for analyzing model behavior across perception, dynamics, and reasoning. 
Solving these tasks requires comprehensive capabilities, including:

\noindent{\textbf{Collision anticipation}} evaluates the capability to reason about \textit{object arrangement and conflicts}. 
In this task, models must predict: (1) whether a collision would occur if a target object moves forward, and (2) whether the collision outcome would change if another specific object were removed from the scene. 
Solving this task demands precise recognition of the spatial attributes of the target object (\eg, orientation, size), and the spatial relationship among different objects (\eg, distance, position).

\noindent{\textbf{Occlusion anticipation}} evaluates the capability to reason about \textit{how object repositioning alters the visible scene structure}.
In this task, objects are arranged in a 3D scene with complex depth relationships. Models must predict how moving a target object affects the visibility of other objects. Success requires understanding the camera perspective, analyzing spatial occlusion relationships between objects, and predicting how spatial adjustments affect visibility.

\noindent{\textbf{Trajectory anticipation}} evaluates the ability to predict object motion under \textit{physical principles and causal constraints}.
In this task, target objects move according to basic physical laws such as gravity, bouncing, and momentum transfer. Models must predict how objects interact with the environment (e.g., whether a moving billiard ball will collide with others and where it will travel).

\noindent{\textbf{Compatibility anticipation}} evaluates the capability to reason about \textit{geometric containment and spatial constraints}.
In this task, models are given a hypothetical placement (\eg, ``If I want to put an apple into the bowl...''), and are required to assess containment feasibility. Specifically, models must determine whether issues such as overflow, spatial overlap, geometric penetration, or physical instability would occur. Success requires fine-grained understanding of object geometry, including size, shape, and volume relationships.

Moreover, for each category, we define two difficulty levels: \textbf{(L1) One-to-one reasoning}, which focuses on reasoning about interactions between a single pair of objects, and \textbf{(L2) One-to-multiple reasoning}, which requires comprehending how a target object simultaneously relates to multiple others within the scene. 
This design enables systematic evaluation of both basic pairwise reasoning and more complex interactions involving multiple objects.

\subsection{Data collection and annotation}
\label{benchmark:collection}
We construct \bench using realistic 3D scenes and assets from publicly available platforms (e.g., CGTrader), where each asset is under its own license specified by the creator and not redistributed. Specifically, we curate a diverse collection of 3D object assets spanning over 50 object categories. To ensure visual realism and diversity, we manually refine the texture of each object asset. Each scene is accompanied by thorough annotations, including RGB images, depth maps, and oriented bounding boxes, which enable accurate question generation. These assets are used solely for rendering purposes and are not redistributed in their original form.

For each scene, we design multiple action prompts that specify hypothetical or counterfactual object motions. These motion events are physically simulated and rendered using Blender's physics engine, ensuring all outcomes adhere to real-world dynamics and physical plausibility. Questions are generated by a set of templates, and Ground-truth (GT) answers are first generated by these simulated outcomes, then manually validated by three individual annotators. Each evaluation instance undergoes rigorous human verification to ensure causal correctness and quality. 

To ensure an unbiased evaluation, we curate an additional real-world validation set. Images are collected from public datasets (e.g., COCO~\cite{lin2014microsoft}, SUN-RGBD~\cite{song2015sun}, and GQA~\cite{hudson2019gqa}), released by CC BY-SA license. The question–answer (QA) pairs are manually annotated. Initial QA proposals generated by one annotator are independently verified by three others. Any disagreements are resolved through majority voting by two additional annotators. Full details are provided in App. Sec.~\ref{sec:app-data}.

\section{Benchmark evaluation}
\label{experiment}
In this section, we detail the experimental setup in Sec.~\ref{experiment:setup}, followed by quantitative results and key observations on \bench in Sec.~\ref{experiment:analysis}. We further analyze failure cases and discuss potential improvement directions in Sec.~\ref{experiment:failures}. 
Our analysis reveals a substantial gap between current MLLMs and human performance, and suggests that this gap may stem from models’ reliance on textual reasoning that is not consistently grounded in real 3D configurations.

\begin{table*}[!t]
\centering
\caption{\textbf{The performance of MLLMs on \bench.} \textit{Col.} denotes Collision anticipation. \textit{Com.} denotes Compatibility anticipation. \textit{Occ.} denotes Occlusion anticipation. \textit{Traj.} denotes Trajectory anticipation. \textit{Avg.} denotes Average scores. \colorbox[HTML]{7FA5D6}{Best} \& \colorbox[HTML]{C3D9E8}{Second best}. The \textcolor{gray}{gray figures} denote the Not Sure Rate (NSR). A clear performance gap between humans and MLLMs.}
\vspace{1em}
\resizebox{\textwidth}{!}{
\begin{tabular}{l|ccccc|ccccc|c|c}
        \toprule
        \multirow{2}{*}{\textbf{Model}} & \multicolumn{5}{c|}{\textbf{Level 1}} & \multicolumn{5}{c|}{\textbf{Level 2}}& \multirow{2}{*}{\textbf{Avg.}} & \multirow{2}{*}{\textbf{Rank}} \\
         & Col. & Com. & Occ. & Traj. & Avg. & Col. & Com. & Occ. & Traj. & Avg. & \\
        \toprule
        \multicolumn{12}{l}{\textbf{Spatial Reasoning MLLMs}} \\
        \midrule
        Spatial-VLM\cite{chen2024spatialvlm} & 20.98 & 47.62 & 48.15 & 46.88 & 36.24 \gray{(3.83)} & 15.75 & 41.67 & 24.07 & 37.96 & 28.25 \gray{(7.17)} & 33.50 & 15\\
        Spatial-MLLM\cite{wu2025spatial} & 41.96 & 55.56 & 48.15 & 54.17 & 47.93 \gray{(1.91)} & 24.41 & 44.44 & 12.96 & 24.82 & 24.56 \underline{\gray{(0.72)}} & 40.00 & 12\\
        \midrule
        \multicolumn{12}{l}{\textbf{Open-source MLLMs}} \\
        \midrule
        LLaVAOneVision1.5-8B-I. \cite{li2024llava} & 43.01 & 42.86 & 38.52 & 29.17 & 38.17 \gray{(15.91)} & 15.38 & 41.67 & 20.37 & 38.69 & 27.84 \gray{(23.30)} & 34.62 & 14\\
        Qwen2.5-VL-7B-I.\cite{qwen3vl} & 41.75 & 42.86 & 42.22 & 44.27 & 42.66 \gray{(2.21)} & 30.40 & 41.67 & 24.07 & 38.69 & 33.79 \gray{(5.73)} & 39.61 & 13\\
        Qwen3-VL-2B-I.\cite{qwen3vl} & 47.43 & 47.62 & 49.63 & 43.23 & 46.69 \gray{(0.88)} & 19.35 & 38.89 & 22.22 & 37.96 & 28.98 \gray{(1.79)} & 40.60 & 11\\
        Qwen3-VL-4B-I.\cite{qwen3vl} & 47.55 & 50.79 & 48.15 & 44.79 & 47.19 \gray{(6.92)} & 24.41 & 44.44 & 16.67 & 62.77 & 40.11 \gray{(10.39)} & 44.76 & 8\\
        Qwen3-VL-8B-I.\cite{qwen3vl} & 41.40 & \cellcolor[HTML]{7FA5D6}66.67 & 51.11 & 47.92 & 47.55 \gray{(2.95)} & 22.76 & 36.11 & 29.63 & 50.36 & 35.85 \gray{(7.89)} & 43.53 & 9\\
        Qwen3-VL-8B-T.\cite{qwen3vl} & 42.74 & \cellcolor[HTML]{C3D9E8}63.49 & 48.15 & \cellcolor[HTML]{C3D9E8}55.73 & {49.43} \gray{(1.03)} & 23.75 & 44.44 & 22.22 & 34.31 & 29.71 \gray{(1.79)} & 42.65 & 10\\
        Qwen3-VL-30B-A3B-I.\cite{qwen3vl} & 47.02 & 55.56 & 48.15 & 47.40 & 48.15 \gray{(1.47)} & 30.16 & 52.78 & 31.48 & 52.55 & 41.32 \gray{(1.08)} & 45.80 & 6\\
        Qwen3-VL-30B-A3B-T.\cite{qwen3vl} &\cellcolor[HTML]{C3D9E8} 52.99 & 58.73 & 51.11 & 54.17 &\colorbox[HTML]{C3D9E8}{53.49} \textcolor[HTML]{374973}{\textbf{(0.00)}} & 35.09 &\cellcolor[HTML]{7FA5D6} 63.89 & 46.30 & 48.18 & 44.79 \textcolor[HTML]{374973}{\textbf{(0.36)}} & 50.50 & 3\\
        % Qwen3-VL-235B-A22B-I.\cite{qwen3vl} & & & & &  & & & & \\
        \midrule
        \multicolumn{11}{l}{\textbf{Closed-source MLLMs}} \\
        \midrule
        GPT5\cite{gpt5} & 49.65 & 61.90 &\cellcolor[HTML]{C3D9E8} 59.26 & 54.17 & \colorbox[HTML]{7FA5D6}{54.00} \gray{(1.47)} & \cellcolor[HTML]{C3D9E8}40.94 & 50.00 & \cellcolor[HTML]{7FA5D6}62.96 & 64.96 & \colorbox[HTML]{7FA5D6}{54.52} \gray{(1.08)} & \cellcolor[HTML]{7FA5D6}54.17 & 1\\
        GPT5-mini\cite{gpt5} & 37.76 & 42.86 & 48.15 & 44.79 & 42.31 \gray{(6.19)} & 35.43 & 47.22 & \cellcolor[HTML]{C3D9E8}50.00 & \cellcolor[HTML]{7FA5D6}67.88 & \colorbox[HTML]{C3D9E8}{51.41} \gray{(3.94)} & 45.44 & 7\\
        Gemini2.5 Pro\cite{gemini-25-pro} & 43.62 & \cellcolor[HTML]{C3D9E8}63.49 & \cellcolor[HTML]{7FA5D6}60.00 & 53.12 & 51.44 \textcolor[HTML]{374973}{\textbf{(0.00)}} & 30.65 & 52.78 & 48.15 & \cellcolor[HTML]{C3D9E8}64.96 & {48.85} \textcolor[HTML]{374973}{\textbf{(0.36)}} &  \cellcolor[HTML]{C3D9E8}50.55 & 2\\
        Gemini2.5 Flash\cite{gemini2.5flash} & 46.21 & 50.79 & 48.89 & \cellcolor[HTML]{7FA5D6}56.25 & 50.02 \underline{\gray{(0.15)}} & \cellcolor[HTML]{7FA5D6}42.31 & \cellcolor[HTML]{C3D9E8}58.33 & 25.93 & 55.47 & 46.53 \gray{(1.79)} & 48.82 & 4\\
        Sonnet 3.7\cite{claude-37-sonnet} & \cellcolor[HTML]{7FA5D6} 55.00 & 49.21 & 45.56 & 46.35 & 50.12 \gray{(5.04)} & 29.41 & 48.61 & 34.26 & 60.58 & 44.16 \gray{(17.29)} & 48.07 & 5 \\
        % Gemini2.0 Flash\cite{gemini2.5flash} & & & & &  & & & & \\
        \midrule
        Human & 86.84 & 80.00 & 92.50 & 82.50 & 85.44 & 78.57 & 97.22 & 97.50 & 62.50 & 83.54 & 84.49 &-\\
        Random &23.61 & 31.75 & 32.35 & 38.54 &30.34 &8.54 &33.33 & 20.37 &32.14 & 24.50 & 28.6 & -\\
        \bottomrule
    \end{tabular}}
    \label{tab:results-on-bench}
\end{table*}

\subsection{Evaluation setup}
\label{experiment:setup}
\noindent \textbf{Evaluated MLLMs.} Our evaluation spans a diverse set of state-of-the-art MLLMs, including spatial-experts (Spatial-VLM\cite{chen2024spatialvlm} and Spatial-MLLM\cite{wu2025spatial}), as well as strongest general MLLMs, including Qwen3-VL series\cite{qwen3vl,bai2025qwen2}), GPT-5 series\cite{gpt5}, Gemini 2.5 series\cite{gemini2.5flash}, and Claude\cite{claude-37-sonnet}. 
Additionally, we recruited three independent human evaluators to complete the \bench with the same questions and choices. 
We also report random guessing scores.

\noindent \textbf{Metrics.}
\label{experiment:metric}
Standard binary evaluation often incentivizes hallucinations by rewarding forced guesses over acknowledging uncertainty~\cite{why-language-models-hallucinate}. 
To address this, we adopt two complementary metrics:
(1) \textbf{True Positive Rate (TPR)}, defined as the proportion of correctly answered questions over all samples, where ``Not Sure'' responses are treated as incorrect; and 
(2) \textbf{Not Sure Rate (NSR)}, defined as the proportion of responses in which the model selects the ``Not Sure'' option.
This dual-metric setup allows us to separate correctness from uncertainty, distinguishing confident errors (hallucinations) from calibrated abstention, and providing a clearer view of model behavior.

\noindent \textbf{Evaluation details.}
\label{experiment:detail}
The evaluation items in \bench are formatted as multiple-choice questions. 
To better understand model reasoning behavior, we prompt models to generate reasoning traces and output their responses in a \{``Reasoning'':..., ``Answer'':...\} JSON format. 
The maximum output length is set to 8192 tokens to allow sufficiently long reasoning traces. 
Each question is evaluated five times to mitigate randomness. All experiments are conducted on two NVIDIA A100 GPUs. 
More details are provided in App. Sec.~\ref{sec:app-details}.

\subsection{Result analysis}
\label{experiment:analysis}

\noindent \textbf{Main results.} 
The results demonstrate a critical disparity between MLLMs and humans.
As detailed in Tab.~\ref{tab:results-on-bench}, humans maintain high accuracy across Level 1 (85.44\%) and Level 2 (83.54\%).
However, MLLMs struggle to bridge this gap.
Even the state-of-the-art GPT5 only achieves 54.17\%, indicating causal spatial reasoning remains a significant challenge for current models.

In the open-source landscape, the Qwen3-VL series demonstrates superior efficiency. 
Qwen3-VL-30B-A3B-T shows the best performance (53.49\% on Level 1 and 50.50\% on Level 2), narrowing the gap with closed-source MLLMs.
The compact Qwen3-VL-2B-I notably outperforms significantly larger models from other families, such as Qwen2.5-VL-7B (39.61\%) and LLaVA-OneVision-8B (34.62\%). 
Furthermore, specialized spatial reasoning MLLMs generally underperform compared to general-purpose MLLMs, highlighting the unique difficulty of our causal spatial tasks.

\begin{figure}[!t]
    \centering
    \includegraphics[width=\linewidth]{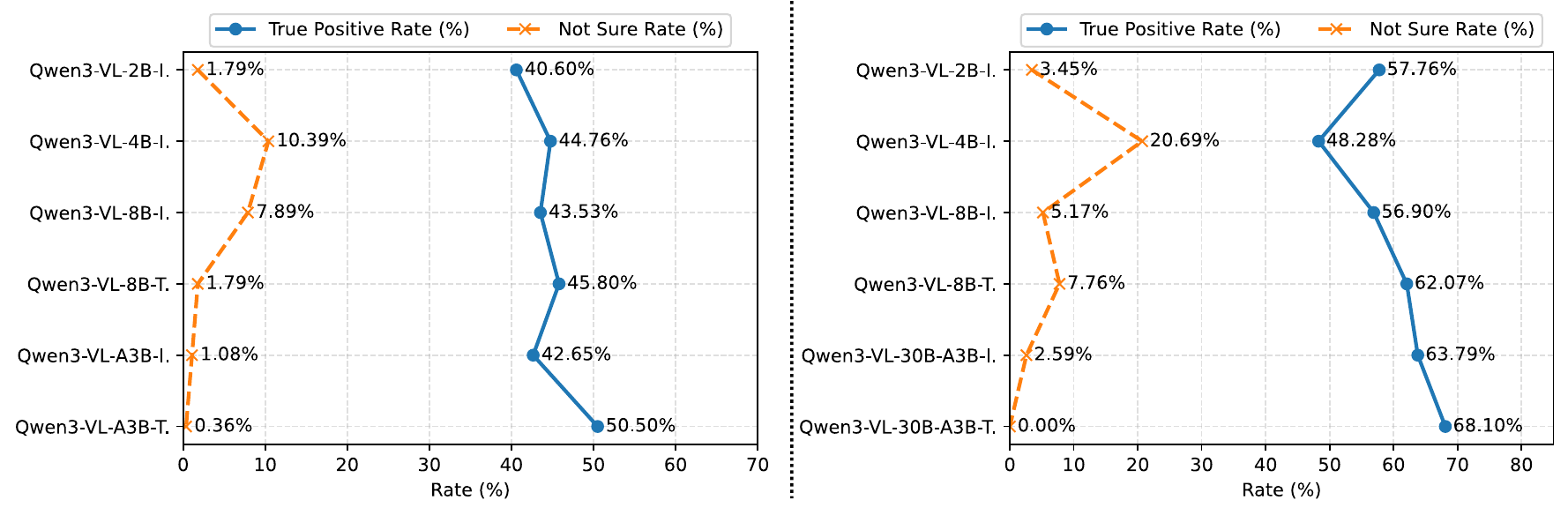}
    \vspace{-7mm}
    \caption{\textbf{Model size analysis on \bench.} Left: synthetic subset; Right: real-world subset. We report True Positive Rate (TPR) and Not Sure Rate (NSR). Larger models tend to be more decisive (lower NSR), while accuracy improvements remain limited.}
    \label{fig:size}
\end{figure}

\noindent \textbf{Not sure rate trends.}
% As shown in Fig.~\ref{fig:size} (left), we observe limited returns from model scaling within the Qwen3-VL series for reasoning within synthetic settings. 
% The instruction-tuned variants exhibit performance plateaus, with the 4B, 8B, and A3B models achieving similar scores of 44.76\%, 43.53\%, and 42.65\%, respectively. 
As shown in Fig.~\ref{fig:size}, we observe a decoupling between model confidence and accuracy. 
For synthetic part in Fig.~\ref{fig:size} (left), in general, larger models exhibit lower Not Sure Rates (NSR), indicating increased decisiveness. 
This trend is further amplified by the thinking mode, where NSR decreases more aggressively (e.g., from 7.89\% (I) to 1.79\% (T) on Qwen3-VL-8B, and from 1.08\% (I) to 0.36\% (T) on Qwen3-VL-A3B). 
Interestingly, the 2B model also shows a relatively low NSR (1.79\%) despite weaker performance, suggesting a tendency toward over-confidence. 
However, increased decisiveness is not consistently accompanied by improved performance. While NSR drops significantly, the corresponding gains in TPR remain limited, indicating that higher confidence does not necessarily translate to better accuracy.

\noindent \textbf{Extension to real-world subset.}
We further evaluate the models on a real-world validation set with more authentic and noisy visual inputs. Results are shown in Fig.~\ref{fig:size} (right). 
Human evaluators achieve 93.10\% average TPR, suggesting that \bench tasks remain well-defined and solvable in real scenarios. 
However, the leading model, Qwen3-30B-A3B-T, falls short by 25\%. 
This persistent gap indicates that the challenges identified in synthetic settings transfer to real-world environments.

Similarly, we observe that larger models tend to exhibit lower uncertainty. 
Qwen3-VL-4B-I shows the highest uncertainty with a Not Sure Rate of 20.69\%, while uncertainty decreases as model size increases. 
The strongest model, Qwen3-VL-30B-A3B-T, reaches 0\%, indicating that larger models tend to produce more decisive outputs even in complex real-world settings, although their accuracy still lags behind human performance.

\begin{figure*}[!t]
    \centering
    \includegraphics[width=1.\linewidth]{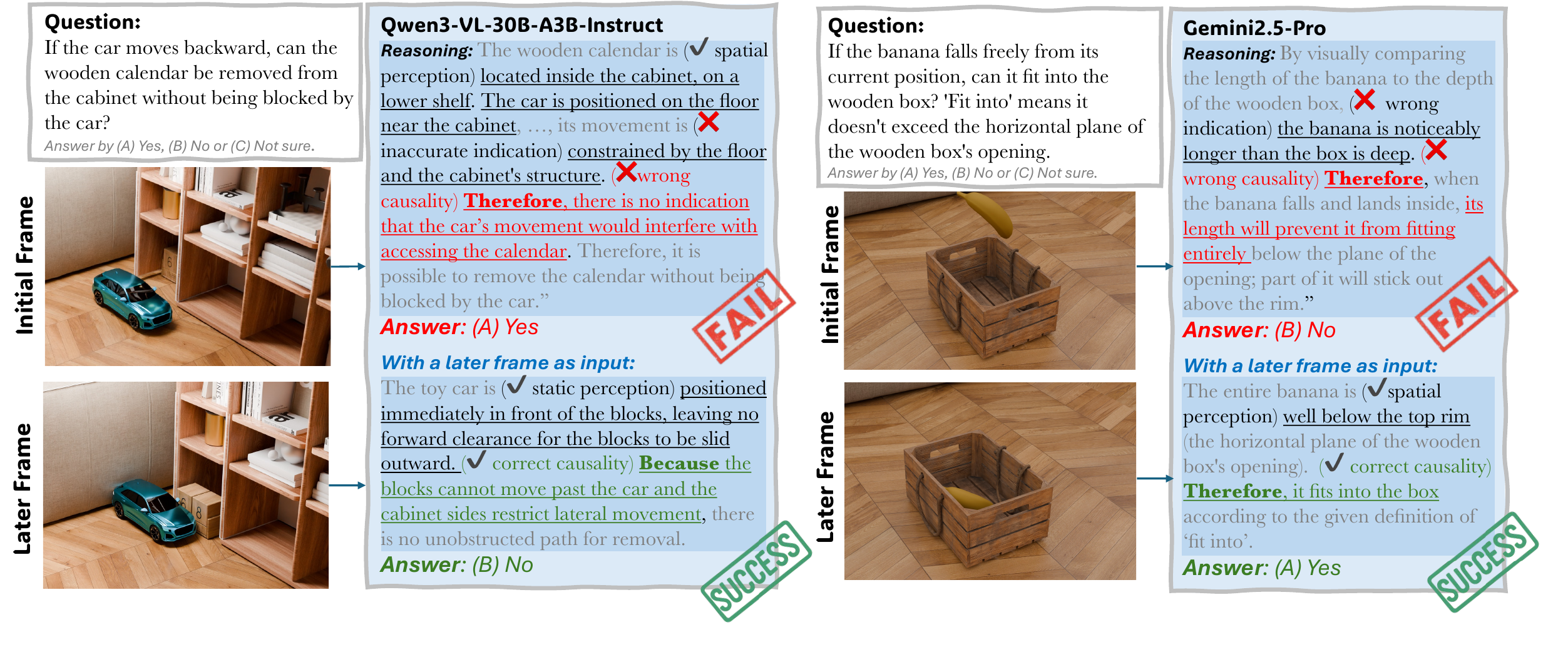}
    \vspace{-10mm}
    \caption{\textbf{Failure cases of MLLMs on \bench.} 
Models often generate fluent but incorrect explanations that are not grounded in visual evidence. 
In both examples, the models rely on textual reasoning and fail to capture the underlying spatial constraints.}
    \label{fig:failures}
    \vspace{-3mm}
\end{figure*}

\subsection{Failure analysis \& improve directions}
\label{experiment:failures}
Despite the challenging nature of causal spatial reasoning, understanding \textit{why} current MLLMs fail and \textit{how} to address these failures provides valuable insights for future progress. We conduct a detailed failure analysis and discuss directions for improvement in this section.

\noindent \textbf{Why do SoTA MLLMs fail in \bench?} 
Figure \ref{fig:failures} illustrates representative failure cases from Qwen3-VL-30B-A3B-Instruct~\cite{qwen3vl} and Gemini2.5-Pro~\cite{gemini-25-pro}. 
Initially, MLLMs generally succeed at identifying static object positions and spatial relationships.
However, a critical disconnect occurs during the subsequent reasoning phase.
These visual cues are either diluted or misinterpreted as the textual chain progresses. In the car–cabinet example, the model produces a coherent explanation but does not consider the car’s visible position in front of the wooden calendar, resulting in an incorrect answer. In the banana–box example, the model concludes that the banana is longer than the depth of the box, which is not supported by the image. We find that this is a common problem for \bench, as illustrated by the quantitative analysis in Appendix~\ref{sec:app-results}.

Consistent with prior studies~\cite{why-is-spatial-reasoning, why-language-models-hallucinate}, we observe that MLLMs prefer to reason based on linguistic priors rather than grounding their logic in the image. 
Consequently, when conducting causal spatial reasoning, models tend to drift toward generic textual CoT, which includes remarkable hallucinations, neglecting the specific geometric constraints present in the visual input. 

This analysis reveals a critical challenge: 

\textit{MLLMs fail to ground their reasoning in the real geometry evidence, drifting toward linguistically plausible but spatially inaccurate explanations with inaccurate prediction.}

\noindent \textbf{How to anchor reasoning in 3D constraints?} 
To solve this visual-grounding related problem, we propose to use World Models (WMs), as they are fundamentally designed to capture causality and temporally consistency given hypothetical actions. However, existing world models~\cite{lu2024genex, wan2025wan} are limited to scene exploration, primarily generating static environments with moving viewpoints, and thus overlook object-level causal spatial reasoning.
Accordingly, we argue that:

\textit{A world model maintaining strict object-level spatial consistency can serve as a reliable visual grounding mechanism, enabling MLLMs to perform accurate causal spatial reasoning.}
\begin{figure*}[!t]
    \centering
    % \vspace{-1mm}
    \includegraphics[width=\linewidth]{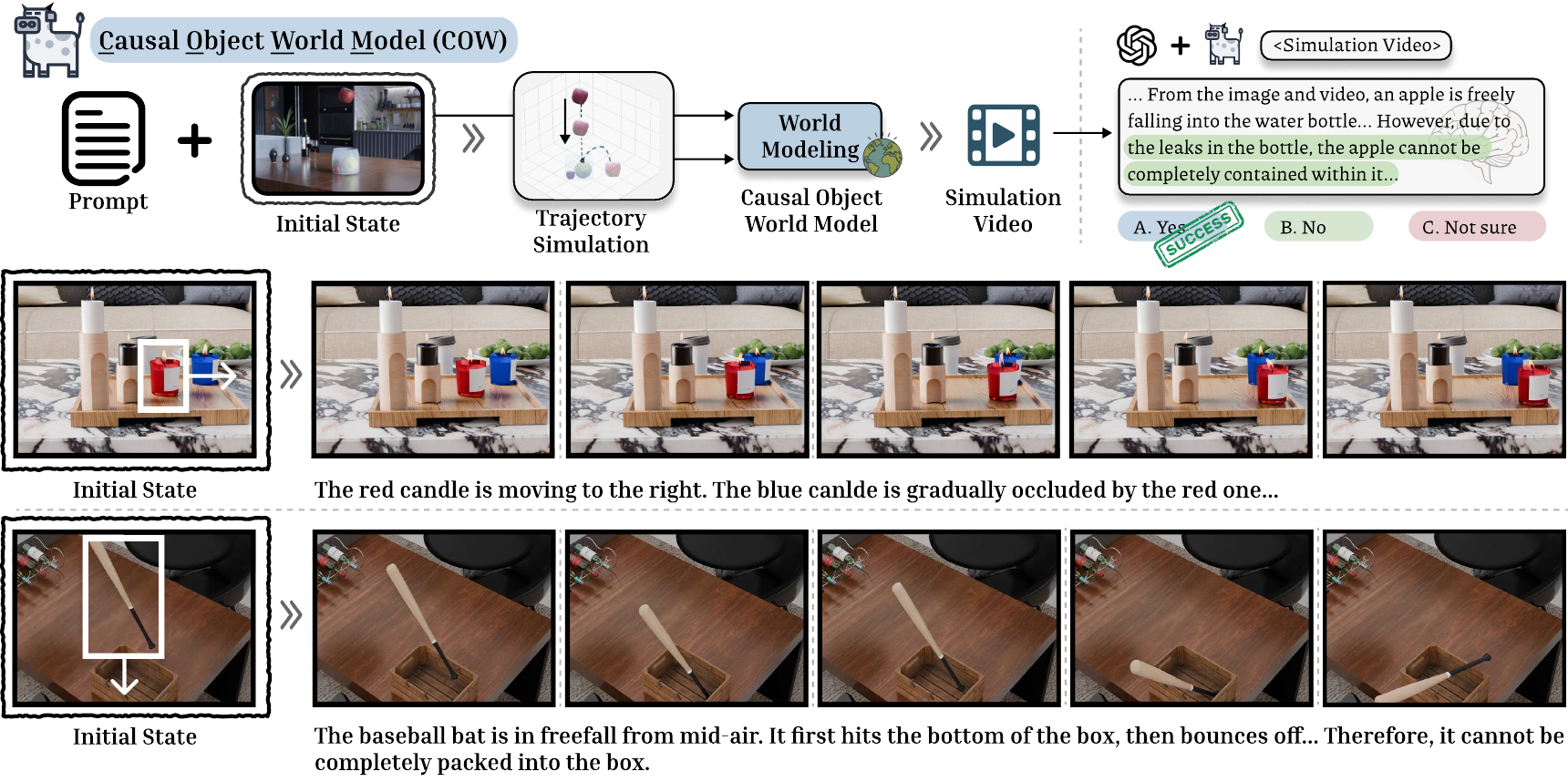}
    \vspace{-5mm}
    \caption{\textbf{(Above) \method pipeline overview}: \method is a 4D trajectory-controlled video generation method that enhances spatial reasoning by explicitly rendering object dynamics through object-centric video generation. \textbf{(Below) Case demonstration}: \method produces physically and visually plausible simulation videos that offer clear visual cues for improved causal spatial inference.
    }
    % \vspace{-3mm}
    \label{fig:pipeline}
\end{figure*}

\section{\method: \Method}

Based on the above discussions, we propose \Method to provide explicit visual cues for causality modeling.
The pipeline design is described in Sec.~\ref{method:method} and depicted in Fig.~\ref{fig:pipeline}.
We evaluate the effectiveness on \bench and discuss the future direction in Sec.~\ref{method:experiment}. Further details are available in App. Sec. \ref{sec:app-iwm}.

\subsection{\method pipeline}
\label{method:method}
The queries from \bench consist of an image representing the initial state and a motion context implying an intervention.
\method processes this information, externalizes causality modeling by trajectory simulation and trajectory-conditioned video generation, providing rich visual cues for MLLM reasoning.
The pipeline includes three key steps. 

\noindent\textbf{4D trajectory simulation.}
We extend the spatial state into a 4D representation using a 6-dimensional vector $Obj_i=\{x, y, z, v_x, v_y, v_z\}$.
The initial position $\{(x, y, z\}$ (predict the bounding box) and $\vec{v}=\{v_x, v_y, v_z\}$ (predict the object orientation or direction indicator in images) are estimated via prompting MLLMs.
We then construct the synthetic trajectory $\tau$ by iteratively updating the object's position, modulated by external physical factors, such as gravity.

\noindent\textbf{Controllable video generation.}
We employ ATI~\cite{wang2025ati} as the world model to render physically consistent dynamics.
The predicted trajectories $\tau_i$ and motion descriptions $m_i$ serve as control signals to condition a diffusion-based generator, synthesizing coherent frames: $F_{later}=\mathcal{W}_\theta(F_{init}|\tau_i^t,d_i)$.
This design uses rich priors of large-scale video models, effectively bridging physical modeling with data-driven imagination for realistic simulation.

\noindent\textbf{Assist MLLMs in reasoning}.
We integrate \method with an MLLM to enable temporally grounded reasoning.
The MLLM jointly interprets the simulated video and the text query to generate informed answers, effectively moving beyond static perception to support motion process prediction.

\subsection{Evaluation with \method}
\label{method:experiment}
\noindent \textbf{Experiment setup.}
We test the representative model, Qwen3-VL-30B-A3B-Instruct.
Our evaluation also includes a subset of Spatial 457\cite{spatial457} that is associated with causal spatial reasoning.
We use three initial frames from generated videos as additional visual cues for subsequent reasoning. 
All images are resized to 512 pixels. 
Experiments are repeated three times to ensure robustness.

% \begin{figure}
%     \centering
%     % \vspace{-1mm}
%     \includegraphics[width=0.55\linewidth]{Figures/Pipeline-Results.pdf}
%     % \vspace{-1mm}
%     \caption{\textbf{Quantative results of \method.} The visual simulation leads to performance gains in \bench, highlighting the potential of explicit vision simulation. The improvement on an external benchmark (Spatial457~\cite{spatial457}) further validates the effectiveness and generalization ability.
%     }
%     \label{fig:iwm-result}
% \end{figure}

\begin{wrapfigure}{tr}{0.55\textwidth}
    % \vspace{-1em}
    \centering
    \includegraphics[width=\linewidth]{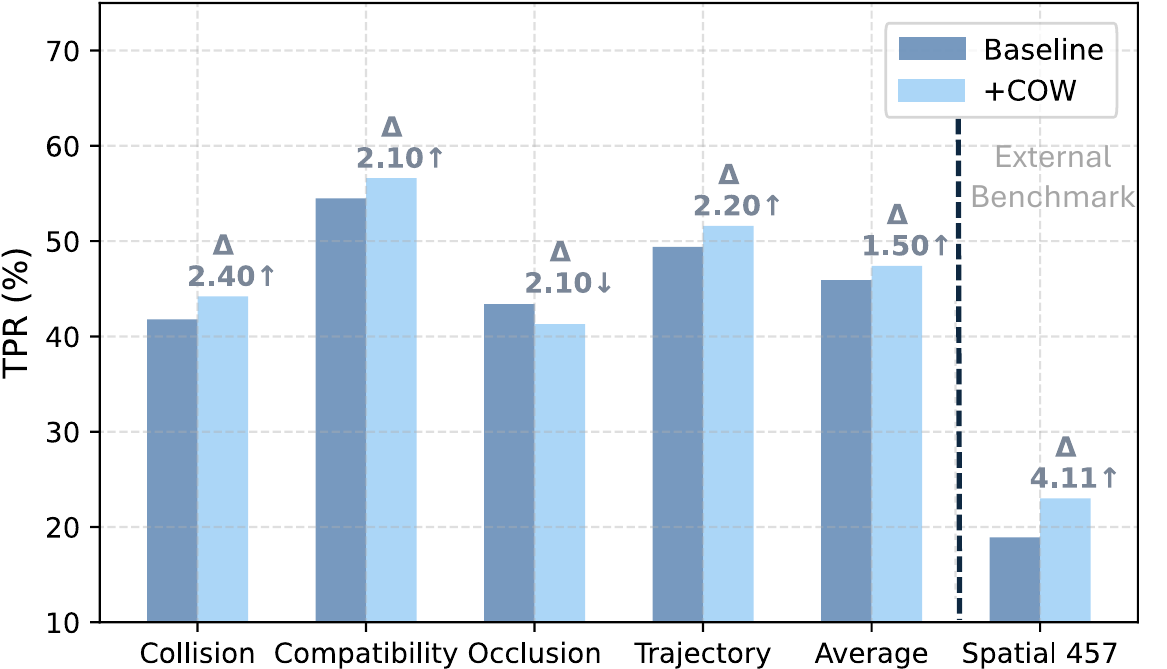}
    % \vspace{-1mm}
    \caption{\textbf{Quantitative results of \method \& Qwen3-30B-A3B-Instruct on \bench and Spatial457~\cite{spatial457}.} 
    The visual simulation leads to performance gains in \bench, highlighting the potential of explicit vision simulation. 
    The improvement on the external benchmark (Spatial457~\cite{spatial457}) further validates the effectiveness and generalization ability.}
    \label{fig:iwm-result}
    % \vspace{-1em}
\end{wrapfigure}

\noindent \textbf{Main results.}
Figure~\ref{fig:iwm-result} demonstrates an average performance boost on \bench. 
Introducing generated video frames yields an accuracy improvement, 2.40\% on Collision and 2.20\% on Trajectory.
Crucially, in our implementation, \method only provides two additional of simulated frames, yet already improves model performance without additional training. 
These results support our hypothesis that grounding reasoning in explicit visual dynamics can improve causal spatial reasoning.
For Spatial457~\cite{spatial457}, which is an external benchmark for evaluation, introducing simulated frames from \method leads to an accuracy improvement of 4.11\%. 
This improvement confirms that \method is effective and generalizable under various data distributions. 

However, there is a slight decrease in the Occlusion task. A possible reason lies in its continuous nature of this task (\eg, an object may be occluded in early frames but reappear later), thus, providing separate simulated frames mainly capture short-term dynamics, providing insufficient cues for long-horizon reasoning. 

\noindent \textbf{Upper bound analysis.}
To further disentangle the role of causal simulation, we replace generated videos with manually validated ground-truth future frames on collision tasks. 
This leads to substantial improvements from baselines (e.g., L1: 47.0 $\rightarrow$ 60.6, L2: 30.2 $\rightarrow$ 52.1), confirming that causal simulation is a key bottleneck. 
However, a considerable gap to human-level performance remains even with perfect future information. Given that causality modeling is already solved in this setting, the results indicate that precise spatial interpretation remains critical and unsolved.

\subsection{Discussion}
\label{sec:cow_discussion}
As illustrated in Fig.~\ref{fig:failures}, simulated future frames help anchor the initial scene and support reasoning about subsequent dynamics. 
In the car-cabinet example, the model correctly identifies the blocking relationship between the car and the cabinet. 
Similarly, in the banana-box example, \method simulates a plausible future state where the banana lies inside the box, enabling correct reasoning about size compatibility. 
These cases suggest that providing explicit future states helps reduce hallucinations by grounding reasoning in visual dynamics rather than relying on linguistic priors. 
This indicates that insufficient causal simulation is a key source of errors in current MLLMs.

However, COW is a training-free probing module rather than a mature end-to-end solution currently. The effectiveness of this approach depends on the quality of the generated simulations. 
In practice, current video generators often lack strict object-level spatial consistency, and downstream models still struggle to interpret fine-grained spatial relations. 
As a result, the improvements observed in Fig.~\ref{fig:iwm-result} show rooms for further improvement.

Nevertheless, achieving consistent gains at test time using only a few simulated frames highlights the value of externalized visual simulation as a lightweight and diagnostic mechanism.
\section{Related Work}
\label{sec:relatedwork}

\textbf{Causal spatial reasoning}.
Recent multimodal advances have empowered spatial reasoning~\cite{cheng2024spatialrgpt,ray2024sat,chen2024spatialvlm, wu2025spatial,qwen3vl,gemini2.5flash, imagine-in-space, sparkle}.
However, existing benchmarks largely remain at the perceptual and simple understanding level.
Early datasets~\cite{clevr,li2023super,hudson2019gqa,tong2024cambrian,liu2023visual, 3dsrbench} focus on \textit{static 3D relations}, while dynamic 4D benchmarks~\cite{li2025sti, zhou2025vlm4d, zhou2025pai, parmar2025chainreaction, lin2025switch} introduce temporal cues but are limited to \textit{passive motion tracking}.
Recent studies~\cite{spatial457, qi2025bear, liu2025beyond, wu2025spatialscore} make more comprehensive efforts, attempting to unify these aspects, yet rely on unrealistic scenes that lack physical fidelity.
Crucially, true spatial intelligence demands \textbf{\textit{causal reasoning}}, not just observing ``what/where is something'', but predicting ``what will happen''. Video-based causal benchmarks such as CausalVQA~\cite{foss2025causalvqa} and VLM4D~\cite{zhou2025vlm4d} provide temporal observations as input, requiring models to perceive dynamics over observed video frames. While valuable, this setting is susceptible to static shortcut biases: as noted in CausalVQA, models can often resolve causal queries by identifying discriminative visual cues in keyframes, without performing genuine causal simulation. 
\bench occupies a distinct position in this landscape. Given only a static image and a textual motion query, models must perform counterfactual causal prediction, reasoning about how a hypothetical intervention transforms the current 3D configuration into a future spatial state. This setting isolates mental simulation from temporal tracking, eliminates shortcut solutions from keyframe cues or temporal correlations, and demands object-grounded causal inference rather than superficial pattern matching. 

\vspace{0.5em}
\noindent\textbf{Video Generation \& World Models.}
World models~\cite{worldmodel, wan2025wan, evoworld, lu2024genex, wang2025ati, emu3-5, mineworld, sora, veo3.1} have shown remarkable success in robotics~\cite{3D-VLA, Zhou2024RoboDreamerLC, ling2025scenethesis, zhang2024combo, long2025survey} and autonomous driving~\cite{hu2023gaia, Gao2024VistaAG, koh2021pathdreamer}, acting as dynamic simulators for anticipation and decision-making~\cite{lecun2022path, seo2023masked, dreamerv2, genie2, wonderworld, wu2023daydreamer}.
However, existing models predominantly focus on global scene dynamics~\cite{alonso2024diffusion, yu2025gamefactory} or ego-motion~\cite{du2023learning, ko2023learning, yang2023learning, zhen2025tesseract}, often lacking the granularity to manipulate specific objects within the scene.
This limitation restricts their utility in complex scenarios where precise object-level interaction is essential.
While recent work has explored object-centric representations~\cite{wu2022slotformer, locatello2020object,didolkar2025ctrl}, these approaches primarily focus on synthetic scenes or image generation and do not extend to the complex physical reasoning required for causal spatial tasks.
To bridge this gap, we introduce \Method (\method).
Unlike holistic scene predictors, \method is an object-centric pipeline specifically targets on trajectory-controlled object motion control, enabling the precise simulation of interactions required for rigorous causal reasoning.
\section{Conclusion}
\label{sec:conclusion}
In this paper, we define a higher-level spatial reasoning paradigm, {Causal Spatial Reasoning}, a significant step to bridge the virtual reasoning and the physical world.
To better investigate this question, we present \bench, the first-of-this-kind benchmark designed to evaluate causal spatial reasoning capabilities. 
Extensive experiments show a substantial gap between MLLMs and human performance, and reveal that this limitation is partly due to reliance on textual reasoning that is not well grounded in visual inputs. 
To address this, we introduced \method, a world-model-based framework that generates controllable object-level motion simulation to support temporally grounded reasoning. 
The experimental improvements highlight the potential of explicit causal imaginations, pointing one possible direction for future exploration.
This research provides the necessary evaluation tools and methodological insights to build AI systems that truly understand the physical laws of the real world, rather than merely matching linguistic patterns.

\section{Acknowledgement}
Alan Yuille was supported by the Office of Naval Research (ONR) under grants N00014-23-1-2641 and N00014-24-1-2696, and National Eye Institute (NEI) with Award ID: R01EY037193. No other grants supported this work. We thank the Johns Hopkins Research IT team (IT@JH) for their support and computing infrastructure.

% \begin{ack}
% Use unnumbered first level headings for the acknowledgments. All acknowledgments

% Do {\bf not} include this section in the anonymized submission, only in the final paper. You can use the \texttt{ack} environment provided in the style file to automatically hide this section in the anonymized submission.
% \end{ack}

\bibliographystyle{plain}
\bibliography{main}

%%%%%%%%%%%%%%%%%%%%%%%%%%%%%%%%%%%%%%%%%%%%%%%%%%%%%%%%%%%%

\appendix
\section{Outline of Appendix}
The Appendix is organized as follows:
\begin{itemize}
    % \item \textbf{Sec.~\ref{sec:relatedwork}} presents the thorough related work section.
    % \item \textbf{Sec.~\ref{sec:comparison-tab}} presents the comparison between \bench and existing spatial reasoning benchmarks, highlighting the unique design of \bench.
    \item \textbf{Sec.~\ref{sec:app-data}} presents the details of our data collection and annotation process. We provide examples of each question template and examples of  \{Image, Question, Answer\} triplets of each reasoning type in this section. Detailed human annotation instructions are provided here.
    \item \textbf{Sec.~\ref{sec:app-details}} presents the details of the evaluation protocol.
    \item \textbf{Sec.~\ref{sec:app-results}} presents additional results analysis. We present a full comparison with standard deviations and reasoning analysis in this section. We include shortcut analysis in this section. Furthermore, quantitative failure analysis is presented here.
    \item \textbf{Sec.~\ref{sec:app-iwm}} presents the additional details of \method.
    \item \textbf{Sec.~\ref{sec:app-iwm-exp}} presents the additional experiment results and analysis of \method. We present visualization, failure case analysis, and additional discussion in this section. 
    \item \textbf{Sec.~\ref{sec:examples}} presents the examples from each category in our benchmark.    
\end{itemize}

\section{Benchmark Data Collection}
\label{sec:app-data}
To comprehensively evaluate the causal spatial reasoning capabilities of MLLMs, \bench is constructed upon fully annotated 3D scene structures that include RGB images, depth maps, and oriented bounding boxes for each object. For every scene, multiple questions are generated with distinct action prompts that specify hypothetical or counterfactual motions. The resulting events are then physically simulated and rendered using physical engines, ensuring that all outcomes are consistent with real-world dynamics and physically plausible interactions.

\subsection{Objects} We collect a diverse set of publicly available 3D object assets spanning multiple semantic categories, including vehicles, decorative items, indoor utilities, sports items (basketball, soccer, and billiards) and everyday objects such as furniture, different containers, and fruits, with more than 50 object types. To enhance visual realism and diversity, we manually refine the textures and curate a consistent color palette for each object. All assets are standardized in scale, material properties, and orientation to ensure coherence across scenes and to enable stable, physically accurate simulations within the rendering environment.  This diversity enables \bench to cover a broad range of spatial interactions, supporting comprehensive evaluation of object-level causal reasoning.

\subsection{Background} To ensure visual realism and physical plausibility, all images in \bench are rendered from fully configured 3D environments. Each scene is composed with carefully arranged backgrounds and lighting conditions to preserve depth cues and contextual coherence. Overall, we sample over 130 diverse camera viewpoints across different elevations and azimuths, capturing rich spatial variability while maintaining consistent scene geometry. This design enables models to perceive spatial layouts from multiple perspectives, supporting robust evaluation of viewpoint-invariant causal reasoning.

\begin{figure}[h]
    \centering
    \includegraphics[width=\linewidth]{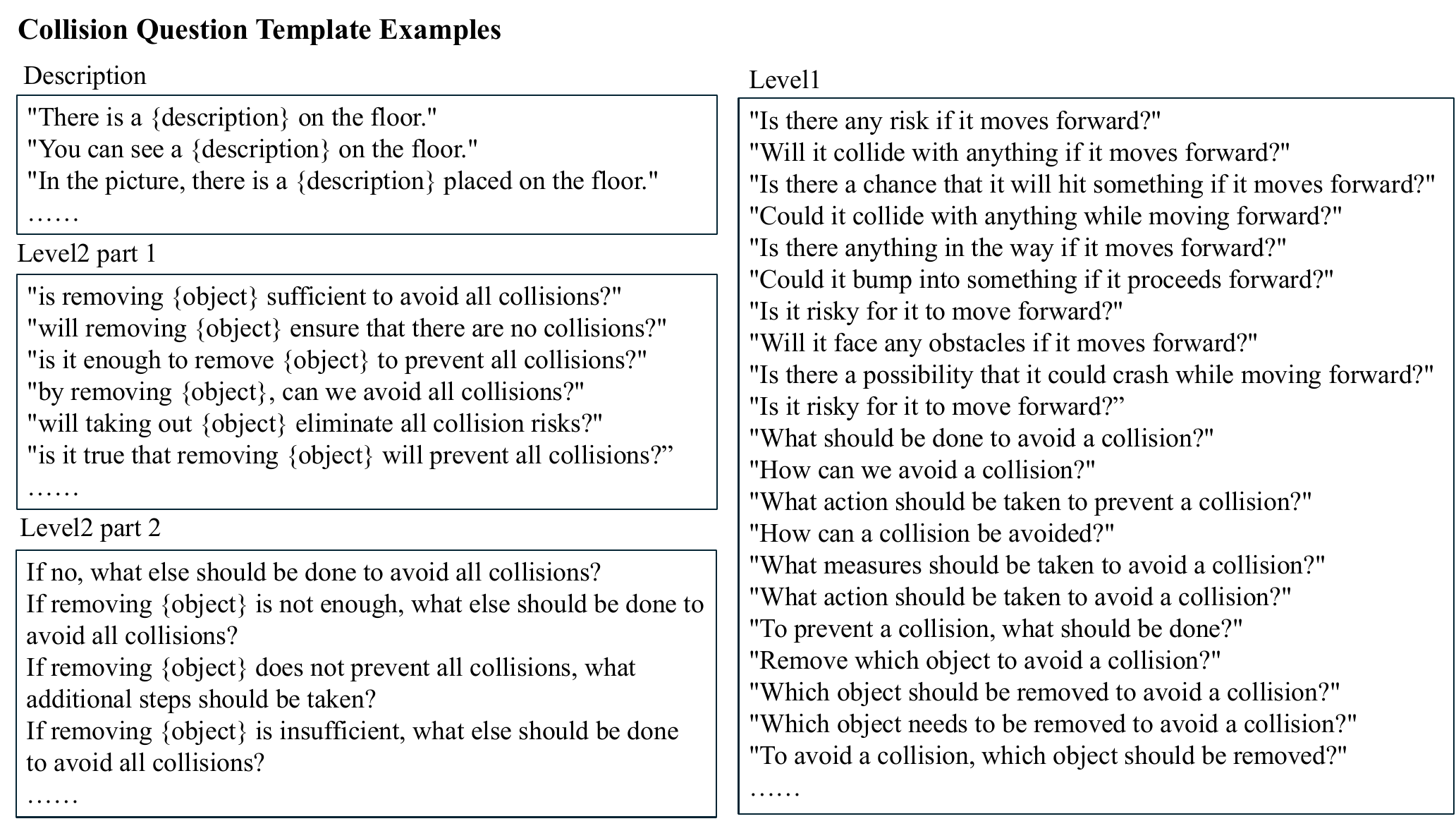}
    \caption{Examples of collision question templates in \bench. Each question is composed by sampled object descriptions and a sampled question either from L1 or L2. The questions from L2 are composed by two parts, with each part sampled separately from ``part1'' and ``part2'' to ensure question diversity. }
    \label{fig:ap-collision-question}
\end{figure}

\begin{figure}[h]
    \centering
    \includegraphics[width=\linewidth]{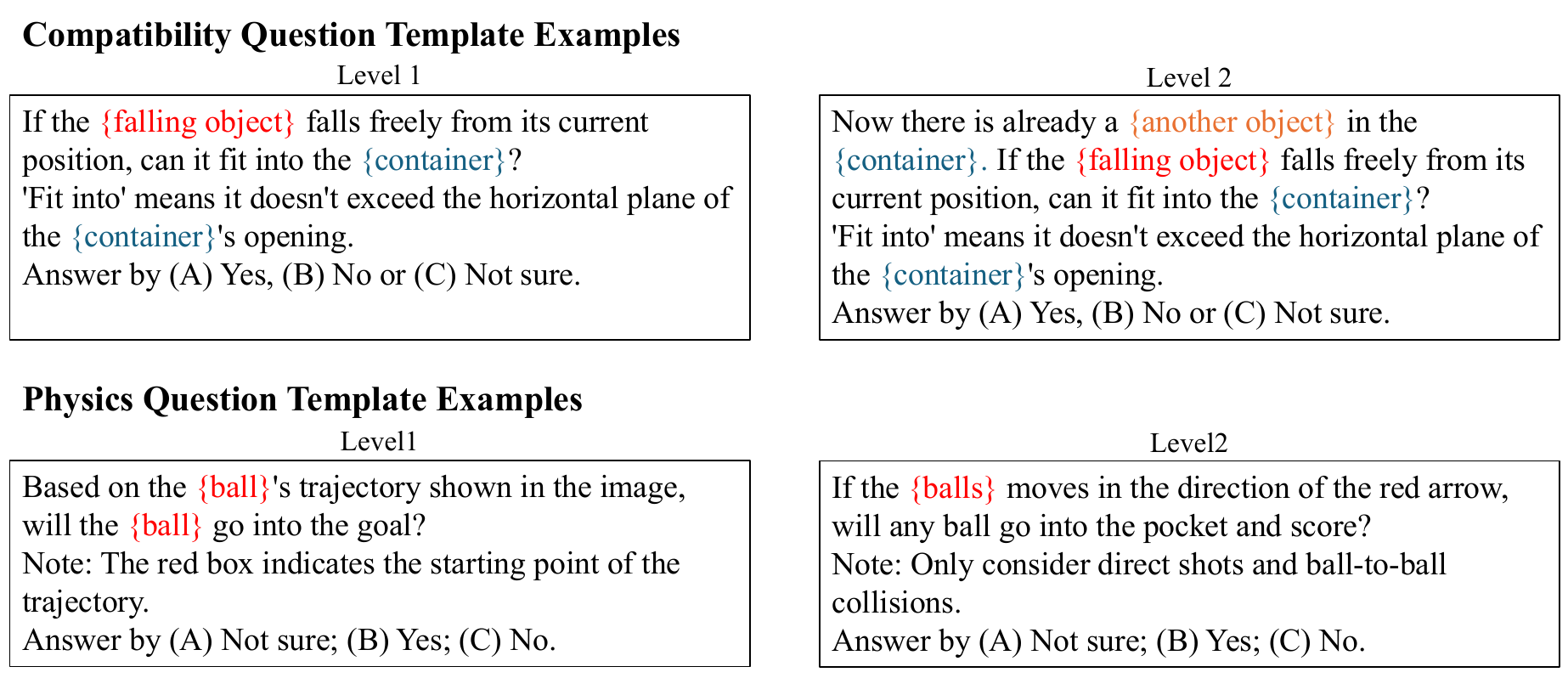}
    \caption{Examples of Compatibility and Trajectory question templates in \bench.}
    \label{fig:ap-com-phy-question}
\end{figure}

For each reasoning type introduced in Sec.~\ref{benchmark:taxonomy}, we design a set of motion prompts ${{M_0, ..., M_k}}$ with explicit directional cues that define object movements. To minimize ambiguity in textual descriptions, motion representations are expressed either as clear textual instructions (for objects with well-defined orientations) or as trajectory arrows (for symmetric objects). GPT-5 is employed to generate and refine object description prototypes for precise object identification, which are subsequently reviewed by human annotators to ensure linguistic clarity and eliminate ambiguity among visually similar objects. A comprehensive set of question templates (as listed in Fig.~\ref{fig:ap-collision-question}, Fig.~\ref{fig:ap-com-phy-question}, and Fig.~\ref{fig:ap-occ-question}) used for benchmark construction is provided in the Appendix. For consistency and ease of evaluation, all questions are formatted as multiple-choice queries with an additional ``not sure'' option included to mitigate random guessing and short-cut biases.

\subsection{Dynamics rendering for answer generation}
After configuring each static scene, we record the center coordinates $(x, y, z)$ and orientation angles $(\text{pitch}, \text{roll}, \text{yaw})$ for every object in the scene. 
After sampling a motion $M_i$, to ensure accurate supervision and physical realism, we simulate each object motion by assigning a constant initial velocity vector $(v_x, v_y, v_z)$ and executing the dynamics within Blender’s Bullet Physics Engine. The simulated scenes are rendered as photo-realistic image sequences that capture realistic trajectories and physically consistent interactions. During this process, the mesh intersection situation is calculated to generate ground truth answer for each question. The first frame of each image sequence is then paired with its corresponding causal reasoning QA to form the final evaluation samples in \bench. 

\subsection{Human Annotation Details}

We recruit three annotators (two PhD students and one research engineer) with backgrounds in computer vision. Prior to the main evaluation, all annotators complete a calibration phase consisting of 20 pilot examples with feedback to align their understanding of the task definitions.
Annotators work independently and have access to a "Not Sure" option, consistent with the model evaluation setting. The average annotation time per question is approximately 1.5 minutes.
In cases of disagreement, two additional annotators are recruited to resolve ambiguities. Final answers are determined by majority vote, requiring agreement from at least three annotators.
For the real-world subset, we adopt a two-stage annotation process: one annotator first proposes question–answer (QA) pairs, which are then verified by three independent annotators. Any disagreement is resolved by two additional annotator for tie-breaking.

\subsection{Statistics}

CausalSpatial uses multiple-choice questions in two formats:
\begin{itemize}
    \item Yes/no-style questions with three options: “Yes,” “No,” and “Not Sure”;
    \item Action/object-selection questions, where the model selects the correct intervention or object-related answer, e.g., which object should be removed to avoid collision or occlusion.
\end{itemize}

Overall, 64.5\% of questions are yes/no-style and 35.5\% require action/object selection. Collision and Occlusion include both formats, with selection questions accounting for 65.4\% and 47.1\%, respectively, while Trajectory and Compatibility are entirely yes/no-style. Most questions have 3 options (68.0\%), with the rest using 4 (19.6\%) or 5 (12.5\%) options. For the real-world subset, 54.3\% are yes/no-style questions and 45.7\% are four-choice descriptive questions.

\section{Additional Evaluation Details}
\label{sec:app-details}

During evaluation, all images are uniformly resized to a resolution of \(1920 \times 1080\) (width \(\times\) height) to preserve the global scene layout and ensure that fine-grained spatial cues (\eg, object boundaries, depth ordering, contact geometry, and subtle occlusion relationships) remain clearly distinguishable to the model. Each benchmark instance is formatted as a multiple-choice question with a varying number of answer options, including a "Not Sure" option to discourage forced predictions and enable uncertainty analysis.

To prevent positional bias and mitigate shortcut exploitation, the ordering of answer choices is independently randomized for every sample. All models are evaluated in a single-turn inference setting with explicit reasoning chain-of-thought prompting.

\section{Additional Results}
\label{sec:app-results}
In this section, we provide an extended analysis of model behavior beyond the main paper results. We examine not only the quantitative differences among various MLLMs, but also the underlying factors that drive these performance gaps across tasks. Together, these results offer a deeper understanding of the capabilities and limitations of current multimodal models in spatial and causal reasoning.

% \begin{figure}
%     \centering
%     \vspace{-2mm}
%     \includegraphics[width=0.97\linewidth]{Figures/Appendix/Radar.pdf}
%     \vspace{-3mm}
%     \caption{\textbf{Holistic Capability Profile.} Values $<$25\% are clamped to the origin. (1) \textbf{Scaling:} Enclosed area expands significantly from Qwen-2B to 30B. (2) \textbf{Balance:} Gemini 2.5 Pro exhibits a convex, balanced shape, contrasting with GPT-5's unbalanced profile (retracted Collision L1 axis).}
%     \label{fig:bench-radar}
%     \vspace{-3mm}
% \end{figure}

\subsection{Comparison Results}
% \wx{Ruisheng jiayou! 1. Spatial MLLMs do not work (methods designed for static 3D reasoning cannot solve in causality); 2. Highlight Qwen3 series; 3. close-source models}

\begin{table*}[t]
\centering
\resizebox{\textwidth}{!}{
\begin{tabular}{l|ccccc|ccccc|c}
        \toprule
        \multirow{2}{*}{\textbf{Model}} & \multicolumn{5}{c|}{\textbf{Level 1}} & \multicolumn{5}{c|}{\textbf{Level 2}}& \multirow{2}{*}{\textbf{Avg.}}  \\
         & Col. & Com. & Occ. & Traj. & Avg. & Col. & Com. & Occ. & Traj. & Avg. \\
        \toprule
        \multicolumn{12}{l}{\textbf{Spatial Reasoning MLLMs}} \\
        \midrule
        Spatial-VLM\cite{chen2024spatialvlm} & 20.98$\pm$1.01 & 47.62$\pm$5.72 & 48.15$\pm$6.11 & 46.88$\pm$2.97 & 36.24$\pm$1.59 & 15.75$\pm$1.92 & 41.67$\pm$1.96 & 24.07$\pm$1.85 & 37.96$\pm$2.58 & 28.25$\pm$1.48 & 33.50$\pm$1.26 \\
        Spatial-MLLM\cite{wu2025spatial} & 41.96$\pm$1.09 & 55.56$\pm$0.00 & 48.15$\pm$0.74 & 54.17$\pm$1.56 & 47.93$\pm$0.21 & 24.41$\pm$1.92 & 44.44$\pm$4.39 & 12.96$\pm$0.00 & 24.82$\pm$1.46 & 24.56$\pm$0.57 & 40.00$\pm$0.07 \\
        \midrule
        \multicolumn{12}{l}{\textbf{Open-source MLLMs}} \\
        \midrule
        LLaVAOneVision1.5-8B-I. \cite{li2024llava} & 43.01$\pm$0.77 & 42.86$\pm$8.09 & 38.52$\pm$1.05 & 29.17$\pm$2.68 & 38.17$\pm$0.38 & 15.38$\pm$3.04 & 41.67$\pm$4.39 & 20.37$\pm$4.72 & 38.69$\pm$1.46 & 27.84$\pm$1.27 & 34.62$\pm$0.63 \\
        DeepSeek-VL2-Tiny$^*$ \cite{wu2024deepseek}  & 25.17 &	34.92 	&22.96 &	43.75 	&30.92 	&14.96 	&27.78 	&20.37 &	26.05 	&21.13 	&27.67 \\
        DeepSeek-VL2-Small$^*$ \cite{wu2024deepseek} & 34.97 &	38.10 	&40.00 	&55.21 	&42.01& 	17.32 	&44.44 	&16.67 	&42.02&28.87 	&37.65 \\
        InternVL3-8B$^*$ \cite{zhu2025internvl3} &	40.91 	&46.03 &	47.41 	&64.58 	&49.41 	&22.05 	&55.56 	&24.07 	&64.71 &	41.07 & 	46.64 \\
        Qwen2.5-VL-7B-I.\cite{qwen3vl} & 41.75$\pm$2.10 & 42.86$\pm$2.51 & 42.22$\pm$2.62 & 44.27$\pm$1.84 & 42.66$\pm$1.16 & 30.40$\pm$6.80 & 41.67$\pm$12.58 & 24.07$\pm$4.14 & 38.69$\pm$9.74 & 33.79$\pm$4.67 & 39.61$\pm$1.18 \\
        Qwen3-VL-2B-I.\cite{qwen3vl} & 47.43$\pm$1.97 & 47.62$\pm$5.61 & 49.63$\pm$3.99 & 43.23$\pm$1.04 & 46.69$\pm$1.04 & 19.35$\pm$6.93 & 38.89$\pm$11.95 & 22.22$\pm$9.44 & 37.96$\pm$5.11 & 28.98$\pm$3.56 & 40.60$\pm$1.62 \\
        Qwen3-VL-4B-I.\cite{qwen3vl} & 47.55$\pm$1.78 & 50.79$\pm$8.09 & 48.15$\pm$2.96 & 44.79$\pm$2.60 & 47.19$\pm$1.25 & 24.41$\pm$6.80 & 44.44$\pm$3.93 & 16.67$\pm$5.86 & 62.77$\pm$1.46 & 40.11$\pm$1.27 & 44.76$\pm$1.19 \\
        Qwen3-VL-8B-I.\cite{qwen3vl} & 41.40$\pm$0.55 & \cellcolor[HTML]{7FA5D6}66.67$\pm$13.70 & 51.11$\pm$2.22 & 47.92$\pm$0.82 & 47.55$\pm$0.96 & 22.76$\pm$3.04 & 36.11$\pm$2.78 & 29.63$\pm$3.70 & 50.36$\pm$4.44 & 35.85$\pm$2.79 & 43.53$\pm$0.53 \\
        Qwen3-VL-8B-T.\cite{qwen3vl} & 42.74$\pm$1.43 & \cellcolor[HTML]{C3D9E8}63.49$\pm$5.02 & 48.15$\pm$2.67 & \cellcolor[HTML]{C3D9E8}55.73$\pm$1.84 &{49.43$\pm$0.23} & 23.75$\pm$3.04 & 44.44$\pm$8.10 & 22.22$\pm$1.85 & 34.31$\pm$0.52 & 29.71$\pm$0.91 & 42.65$\pm$0.15 \\
        Qwen3-VL-30B-A3B-I.\cite{qwen3vl} & 47.02$\pm$1.57 & 55.56$\pm$3.55 & 48.15$\pm$2.62 & 47.40$\pm$3.33 & 48.15$\pm$0.53 & 30.16$\pm$7.32 & 52.78$\pm$11.45 & 31.48$\pm$6.68 & 52.55$\pm$3.26 & 41.32$\pm$1.04 & 45.80$\pm$0.66 \\
        Qwen3-VL-30B-A3B-T.\cite{qwen3vl} &\cellcolor[HTML]{C3D9E8} 52.99$\pm$1.43 & 58.73$\pm$4.05 & 51.11$\pm$2.22 & 54.17$\pm$3.49 &\colorbox[HTML]{C3D9E8}{53.49$\pm$1.13} & 35.09$\pm$1.36 &\cellcolor[HTML]{7FA5D6} 63.89$\pm$12.42 & 46.30$\pm$7.97 & 48.18$\pm$2.13 & 44.79$\pm$3.52 \textcolor[HTML]{374973} & 50.50$\pm$1.82 \\
        % Qwen3-VL-235B-A22B-I.\cite{qwen3vl} & & & & &  & & & & \\
        \midrule
        \multicolumn{11}{l}{\textbf{Closed-source MLLMs}} \\
        \midrule
        GPT5\cite{gpt5} & 49.65$\pm$2.08 & 61.90$\pm$6.35 &\cellcolor[HTML]{C3D9E8} 59.26$\pm$1.48 & 54.17$\pm$5.21 & \colorbox[HTML]{7FA5D6}{54.00$\pm$3.24} & \cellcolor[HTML]{C3D9E8}40.94$\pm$0.00 & 50.00$\pm$0.00 & \cellcolor[HTML]{7FA5D6}62.96$\pm$1.85 & 64.96$\pm$1.46 & \colorbox[HTML]{7FA5D6}{54.52$\pm$0.36}  & \cellcolor[HTML]{7FA5D6}54.17$\pm$2.40 \\
        GPT5-mini\cite{gpt5} & 37.76$\pm$2.08 & 42.86$\pm$2.38 & 48.15$\pm$8.15 & 44.79$\pm$1.04 & 42.31$\pm$2.83  & 35.43$\pm$3.85 & 47.22$\pm$2.78 & \cellcolor[HTML]{C3D9E8}50.00$\pm$3.70 & \cellcolor[HTML]{7FA5D6}67.88$\pm$2.19 & \colorbox[HTML]{C3D9E8}{51.41$\pm$0.64} & 45.44$\pm$2.28 \\
        Gemini2.5 Pro\cite{gemini-25-pro} & 43.62$\pm$3.46 & \cellcolor[HTML]{C3D9E8}63.49$\pm$3.17 & \cellcolor[HTML]{7FA5D6}60.00$\pm$5.19 & 53.12$\pm$2.08 & 51.44$\pm$3.39 & 30.65$\pm$1.92 & 52.78$\pm$2.78 & 48.15$\pm$14.81 & \cellcolor[HTML]{C3D9E8}64.96$\pm$0.73 & {48.85$\pm$3.94} &  \cellcolor[HTML]{C3D9E8}50.55$\pm$1.25 \\
        Gemini2.5 Flash\cite{gemini2.5flash} & 46.21$\pm$0.98 & 50.79$\pm$11.51 & 48.89$\pm$4.10 & \cellcolor[HTML]{7FA5D6}56.25$\pm$2.89 & 50.02$\pm$2.42  & \cellcolor[HTML]{7FA5D6}42.31$\pm$10.04 & \cellcolor[HTML]{C3D9E8}58.33$\pm$2.95 & 25.93$\pm$7.00 & 55.47$\pm$1.83 & 46.53$\pm$1.71  & 48.82$\pm$1.75 \\
        Sonnet 3.7\cite{claude-37-sonnet} & \cellcolor[HTML]{7FA5D6}55.00 $\pm$1.04 & 49.21$\pm$1.59 & 45.56$\pm$6.30 & 46.35$\pm$0.00 & 50.12$\pm$1.84 & 29.41$\pm$0.00 & 48.61$\pm$1.39 & 34.26$\pm$0.93 & 60.58$\pm$0.73 & 44.16$\pm$0.36 & 48.07$\pm$1.41 \\
        % Gemini2.0 Flash\cite{gemini2.5flash} & & & & &  & & & & \\
        \midrule
        Human & 86.84 & 80.00 & 92.50 & 82.50 & 85.44 & 78.57 & 97.22 & 97.50 & 62.50 & 83.54 & 84.49 \\
        Random &23.61 & 31.75 & 32.35 & 38.54 &30.34 &8.54 &33.33 & 20.37 &32.14 & 24.50 & 28.6 \\
        \bottomrule
    \end{tabular}}
    \caption{\textbf{The performance of MLLMs on \bench.} \textit{Col.} denotes Collision. \textit{Com.} denotes Compatibility. \textit{Occ.} denotes Occlusion. \textit{Traj.} denotes Trajectory. \textit{Avg.} denotes Average scores. \colorbox[HTML]{7FA5D6}{Best} \& \colorbox[HTML]{C3D9E8}{Second best}.}
    \label{tab:results-on-bench-suppl}
    \vspace{-2mm}
\end{table*}

In Tab.~\ref{tab:results-on-bench-suppl}, we present the performance of various MLLMs across diverse causal spatial tasks and difficulties on \bench. We compare proprietary models (\eg, GPT-5, Gemini 2.5 Pro), open-source general models (\eg, Qwen3-VL series, LLaVA-OneVision), and open-source spatial intelligence models (\eg, Spatial-VLM, Spatial-MLLM) to assess their causal spatial reasoning capabilities.

\textbf{Comparison across diverse model groups.} The results reveal distinct performance tiers among the evaluated groups. Closed-source frontier models generally lead the benchmark, with GPT-5 achieving a strong average score of 54.17\%, followed closely by Gemini 2.5 Pro at 50.55\%. Top-tier open-source models have largely closed the gap. Conversely, spatially-tuned models struggle, with Spatial-VLM and Spatial-MLLM averaging only 33.50\% and 40.00\%. This indicates that current spatial tuning prioritizes static perception over the causal spatial reasoning demanded by our benchmark. Furthermore, despite the advancements in general MLLMs, a substantial gap remains between the best model performance ($\sim$55\%) and human performance (84.49\%), highlighting the challenging nature of causal spatial reasoning.

\textbf{Comparison in Qwen series.} The Qwen3-VL series serves as an excellent testbed for analyzing the effects of model scaling and reasoning mechanisms. As shown in Tab.~\ref{tab:results-on-bench-suppl}, the performance does not follow a strict linear scaling law. While the comprehensive score improves from 2B (40.60\%) to 4B (44.76\%), it exhibits a slight decline for the 8B model (43.53\%), reaching a performance plateau. 
Additionally, extra ``thinking'' mode training fails to lead to consistent performance improvement.
For the 8B model, enabling the thinking mode results in a performance regression (dropping from 43.53\% to 42.65\%), whereas for the 30B model, the thinking mechanism unlocks significant potential, boosting performance from 45.80\% to 50.50\%. 
This contrast indicates that while larger models can effectively leverage extended reasoning for causal spatial tasks, smaller models may struggle to utilize the additional tokens effectively, potentially leading to reasoning errors.

\textbf{Comparison in Proprietary Models.} 
We further examine frontier models to understand the trade-offs between capability balance and model capacity. As detailed in Tab.~\ref{tab:results-on-bench-suppl}, GPT-5 achieves the leading position among proprietary models with an average score of 54.17\%, demonstrating robust performance across diverse spatial tasks. 
Gemini 2.5 Pro follows with a competitive score of 50.55\%, while Claude 3.7 Sonnet records 48.07\%. 
Notably, we observe a sharp performance decline when transitioning to lightweight model variants. For instance, GPT-5 outperforms its mini counterpart by nearly 9 percentage points (54.17\% vs. 45.44\%). This consistent gap confirms that causal spatial reasoning is a capacity-hungry task, where lightweight models lack the sufficient depth to simulate complex physical interactions.

\subsection{Failure Analysis}
To examine why performance shows limited gain despite significantly longer generation from reasoning-oriented models, we analyze the reasoning traces produced on causal spatial tasks. Upon our analysis, MLLMs reliably recognize static layouts and object placements. However, in failure cases, the visual information in the image is either weakly integrated or incorrectly interpreted. When this happens, the model’s reasoning becomes dominated by generic linguistic patterns learned during instruction tuning, rather than by the specific spatial geometry present in the scene.

We conduct a failure analysis using an LLM-as-a-judge paradigm to isolate where errors originate in chain-of-thought (CoT) reasoning. We design four complementary metrics: (1) \textbf{Visual Grounding Accuracy (VG Acc)}: whether the model correctly uses visual evidence from the image without missing critical aspects, (2)\textbf{Textual Dynamic Understanding Accuracy (TDU Acc)}: whether the model correctly understands the textual motion description, (3) \textbf{Cause Understanding Rating (1--3)}: whether the model correctly identifies the causal trigger of the target action, and (4) \textbf{Cause$\rightarrow$Effect Rating (1--3)}: whether the model correctly propagates the causal chain.
To mitigate hallucinations, we cross validate VG Acc by prompting the judge to explicitly identify missing key factors if judge thinks visual grounding process has missed some key factors, which rate is denoted as \textbf{Missing Rate}.

\begin{table}[t]
\centering
\small
\begin{tabular}{lcccccc}
\toprule
\textbf{Model} & \textbf{Avg} & \textbf{VG Acc} & \textbf{Missing Rate} & \textbf{TDU Acc} & \textbf{Cause} & \textbf{Cause$\rightarrow$Effect} \\
\midrule
Qwen3-VL-4B & 44.8 & 2.1\% & 95.8\% & 43.3\% & 1.70 / 3 & 1.57 / 3 \\
Qwen3-VL-30B-A3B & 45.8 & 2.3\% & 93.7\% & 66.4\% & 2.00 / 3 & 1.70 / 3 \\
Gemini2.5-pro & 50.6 & 3.2\% & 91.7\% & 68.1\% & 2.05 / 3 & 1.76 / 3 \\
\bottomrule
\end{tabular}
\caption{Failure analysis across representative MLLMs. VG: Visual Grounding. TDU: Textual Dynamic Understanding.}
\label{tab:failure_analysis}
\end{table}

Across three representative models, we observe a consistent pattern: 
\textbf{Text understanding is not the bottleneck.} Models achieve relatively strong textual dynamic understanding, indicating they can generally parse motion descriptions correctly.
\textbf{In contrast, visual grounding is the primary failure source.} Visual grounding accuracy is extremely low (2--3\%), while over 90\% of errors involve missing key visual factors. This suggests that models largely ignore or hallucinate critical visual evidence.
Causal errors stem from mis-grounded reasoning. Cause Understanding and Cause$\rightarrow$Effect scores remain moderate (approximately 1.7--2.0/3). Although models attempt causal simulation, errors arise at the cause identification stage because reasoning is conditioned on incorrect visual premises, leading to plausible but mis-grounded conclusions.

\subsection{Answer-First Ablation}
To examine whether the observed failures are primarily induced by explicit chain-of-thought (CoT) generation, we additionally evaluate an answer-first variant, where the model is prompted to provide the answer before generating its rationale. As shown in Table~\ref{tab:answer_first}, removing the requirement to reason before answering does not lead to consistent performance improvements. These mixed results suggest that explicit CoT generation itself is not a consistent cause of the observed failures. Instead, our qualitative analysis indicates that models can produce fluent and seemingly coherent rationales even when the underlying visual evidence is incomplete or incorrectly grounded. Therefore, we interpret linguistic hallucination here not as hallucination caused by CoT prompting, but as a manifestation of erroneous or insufficient visual grounding that is subsequently rationalized in language. This ablation thus helps distinguish failures in visual evidence acquisition/grounding from those caused solely by the reasoning format.

\begin{table}[t]
\centering
\caption{Comparison between reasoning-first and answer-first prompting.}
\label{tab:answer_first}
\begin{tabular}{|l|c|c|}
\hline
\textbf{Model} & \textbf{Reasoning-first} & \textbf{Answer-first} \\
\hline
Qwen3-VL-8B-I      & \textbf{43.5 $\pm$ 0.53} & 42.0 $\pm$ 2.38 \\
Qwen3-VL-30B-A3B-I & 45.8 $\pm$ 0.66 & \textbf{48.4 $\pm$ 1.32} \\
Qwen3-VL-32B-I     & \textbf{52.2 $\pm$ 0.60} & 45.2 $\pm$ 1.66 \\
\hline
\end{tabular}
\end{table}

\subsection{Shortcut and Leakage Analysis}

To address potential concerns about whether models can exploit template-level linguistic patterns or answer-option shortcuts rather than performing genuine causal spatial reasoning, we conduct a controlled ablation study with four degraded input conditions:

\begin{itemize}
    \item \textbf{Text-only:} The question image is removed; the model receives only the text prompt.
    \item \textbf{Image-only:} The question text is removed; the model receives only the image.
    \item \textbf{Option-only:} Both image and question text are removed; the model receives only the answer choices.
    \item \textbf{Template-only:} The question-specific text is replaced with a fixed category-level template description, removing instance-specific content.
\end{itemize}

Results are reported in Tab.~\ref{tab:shortcut}, where Not Sure Rates (NSR) are shown in parentheses.

\begin{table}[h]
\centering
\caption{Shortcut ablation results across four degraded input conditions. Numbers in parentheses denote Not Sure Rate (NSR, \%). Default refers to the standard full-input evaluation setting.}
\label{tab:shortcut}
\resizebox{\linewidth}{!}{%
\begin{tabular}{lccccc}
\toprule
\textbf{Model} & \textbf{Text-only} & \textbf{Image-only} & \textbf{Option-only} & \textbf{Template-only} & \textbf{Default} \\
\midrule
Qwen3-VL-4B         & 12.8 \scriptsize{(54.0)} & 27.7 \scriptsize{(8.1)}  & 1.5 \scriptsize{(72.0)}  & 0.7 \scriptsize{(74.5)}  & 44.7 \\
Qwen3-VL-30B-A3B    & 13.2 \scriptsize{(45.8)} & 35.2 \scriptsize{(12.5)} & 14.0 \scriptsize{(65.0)} & 5.5 \scriptsize{(85.1)}  & 50.5 \\
Qwen3-VL-32B        & 9.5  \scriptsize{(64.6)} & 34.9 \scriptsize{(16.6)} & 4.1 \scriptsize{(90.9)}  & 1.1 \scriptsize{(96.4)}  & 52.2 \\
Gemini~2.5~Flash    & 18.2 \scriptsize{(26.9)} & 26.4 \scriptsize{(4.4)}  & 2.8 \scriptsize{(75.4)}  & 2.2 \scriptsize{(67.8)}  & 48.8 \\
\bottomrule
\end{tabular}%
}
\end{table}

\noindent\textbf{Analysis.}
The results provide no evidence of shortcut exploitation or data leakage. Several clear patterns emerge. First, removing either the image (\textit{text-only}) or the question text (\textit{image-only}) causes a large and consistent drop in accuracy relative to the default setting --- from roughly 45--52\% down to 9--35\% --- confirming that both modalities carry complementary and necessary information for solving \bench tasks. Neither alone is sufficient. Second, under the \textit{option-only} and \textit{template-only} conditions, models overwhelmingly default to selecting ``Not Sure'' (NSR of 65--97\%), indicating that answer choices and category-level templates carry little exploitable signal in isolation. Importantly, performance under these conditions collapses well below the random-guess baseline, ruling out systematic answer-bias exploitation. Together, these findings confirm that strong performance on \bench requires genuine integration of visual and textual inputs, and cannot be attributed to pattern matching within question templates or answer options.

\section{Implementation Details of \method}
\label{sec:app-iwm}
\Method (\method) is a modular world model framework.
It can be adopted in the causal spatial reasoning scenarios, such as occlusion, compatibility, collision, and Trajectory.
\method consists of four primary steps: instance grounding, motion analysis, trajectory computation, and video generation.

\subsection{Instance Grounding}
Each question in \bench consists of an image depicting the current spatial state and a question text that implies a specific motion.
At the core of causal spatial reasoning lies the notion of an instance: the entity whose spatial state first undergoes change and subsequently interacts with other entities.
Accordingly, the initial stage of our \method focuses on instance grounding.

To identify the instance involved in the motion, we employ an MLLM (Qwen3-VL-A3B) to interpret the instance that involves the motion described in the question.
For simple cases, for example, the query ``the car is moving forward'', the model can readily determine that ``the car'' is the instance whose state changes.
More complex scenarios require finer-grained grounding.
For example, in the scenario of billiards collisions, the image may contain multiple billiards balls, and the moving ball must be explicitly distinguished (\eg, ``the red number-5 ball'', ``the white ball'', ``the black ball'').
To handle this, we first prompt the MLLM to generate a precise textual description of the moving instance.
Then, given this description, we further prompt the model to localize the pixel coordinate $(x_p, y_p)$ of the corresponding instance within the image.

\subsection{Motion Analysis}
The second step in our pipeline is to analyze the motion described in the question.
Our questions typically involve two distinct categories of moving instances, each requiring a different approach for motion analysis.
The first category includes instance with a clearly defined orientation, such as ``the car''.
In these cases, the query might describe the motion in terms of specific directions, like ``moving forward'' or ``moving to the left''.
To determine the exact direction of motion, we rely on the relative positioning of the instance orientation within the image.
We can infer the concrete direction of movement by analyzing these positional relationships.

The second category involves instances that lack clear orientation, such as billiards balls, vases, or basketballs.
For these instances, the image includes explicit visual cues, such as arrows, to indicate the direction of motion.
Additionally, the motion direction is parallel to the arrow's direction, but the query itself determines whether the arrow should be considered positive or negative. 
We prompt the MLLM to identify the direction of the arrow in the image, and based on the question's context, we assign the appropriate positive or negative value to the motion direction.

\subsection{Trajectory Computation}
After obtaining both the pixel coordinates of the target instance and its motion direction, we proceed to simulate its movement.
We first preset the number of frames and assign a constant velocity.
The position of the instance is defined as the center pixel coordinate of its bounding box in the initial frame.
By combining the motion direction with the preset velocity and frame count, we compute a sequence of center coordinates that represent the instance’s movement over time.
This produces a series of pixel-level positions, forming the trajectory tensors of the moving instance.

\subsection{Video Generation}
After the trajectory tensors are obtained, we feed the detailed motion description, the question image, and the constructed trajectory tensors into the ATI video generation model to synthesize the simulation video.
The question image provides the initial spatial state of the scene.
The motion description supplies additional semantic cues for generating motion-consistent content.
The trajectory tensor explicitly controls the instance’s movement.

Concretely, the input image is first encoded by a VAE encoder into a latent feature map $L_1\in R^{H\times W\times C}$.
For the input trajectory, ATI uses the starting location of the point in the first frame to bilinearly sample a $C$-dimensional feature vector from $L_1$, which serves as the appearance feature associated with that trajectory.

\begin{figure}[!h]
    \centering
    \includegraphics[width=0.98\linewidth]{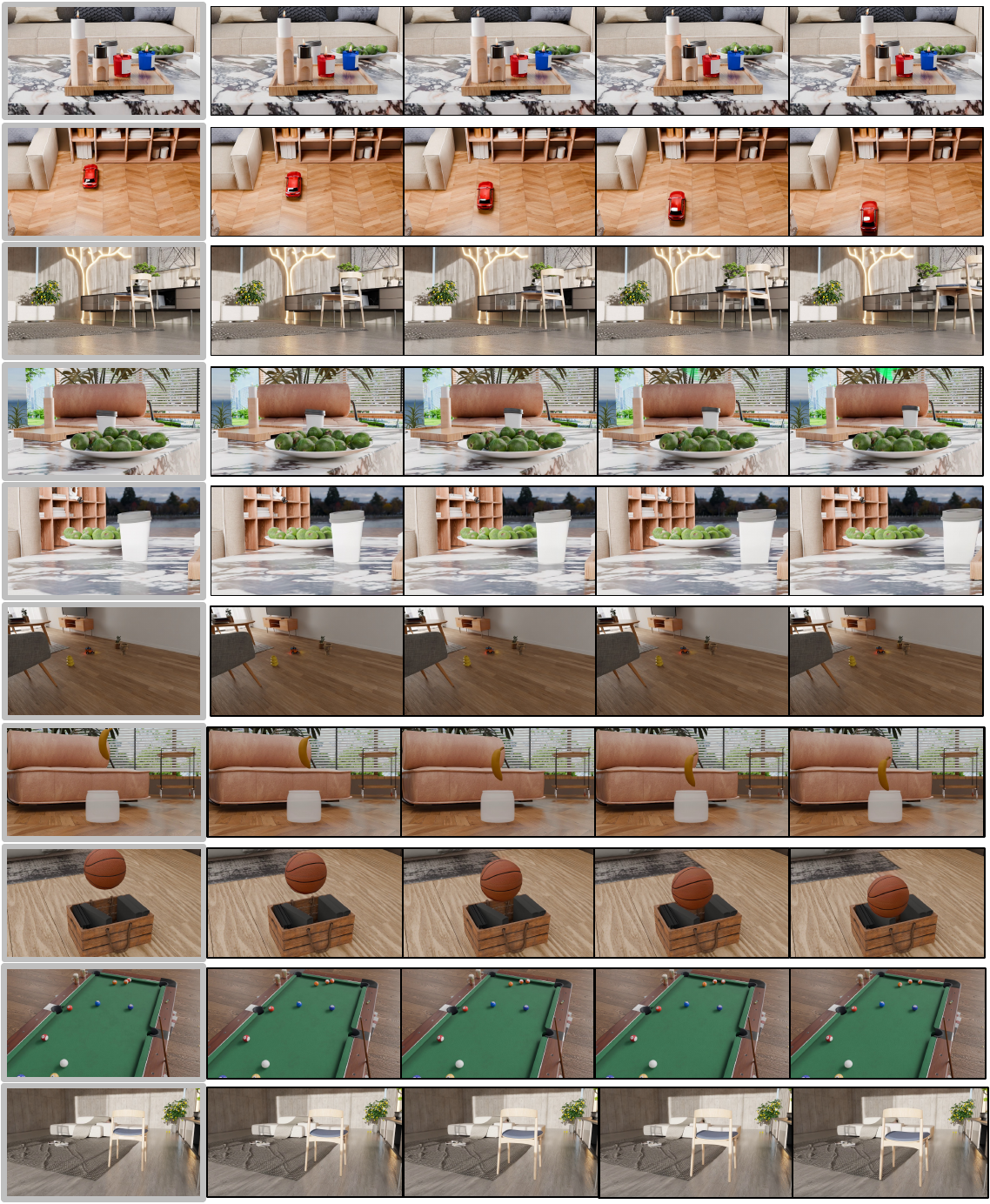}
    \caption{Additional visualizations of \method across diverse indoor and outdoor scenes. Despite the presence of noisy backgrounds with complex object arrangements, \method consistently preserves object boundaries and appearance, maintains stable geometric reasoning, and produces coherent spatial predictions. This explicit visual information helps MLLMs with causal spatial reasoning tasks.}
    \label{fig:ap-visualization}
\end{figure}

Furthermore, we randomly sample a few static points which is away from the motion trajectory, used for fixing the camera view.
Therefore, for each generation, we have a trajectory tensor $\phi _{M}=(x_t, y_t)_{t=0}^T$ and a set of static points $\Phi_{s} = \{\phi_s\}^{N}$, where $\{\phi _{s}=(x_i, y_i)_{t=0}^T\}$.
In total, the input trajectory is $\Phi = \{\phi _M, \Phi_s\}$.
ATI then constructs, at each frame $t$, a spatial Gaussian mask centered at $(x_t, y_t)$ in the latent grid.
This mask softly spreads the sampled feature to nearby latent pixels, assigning larger weights to positions closer to the trajectory point and decaying smoothly with distance.
In this way, the same appearance feature is propagated along the trajectory over time.

These trajectory-conditioned latent features are then injected into the pre-trained image-to-video DiT backbone as an additional conditioning signal. 
During denoising, the DiT blocks jointly process the noisy video latents and the injected trajectory features, guiding the diffusion process to follow the specified motion paths while preserving semantic and visual coherence. Finally, the denoised video latents are decoded by the VAE decoder into RGB frames, producing a video whose motion is aligned with the user-provided trajectories. In practice, ATI can be plugged into different I2V backbones (Wan2.1-14B) without retraining them from scratch, and simple regularization strategies such as tail-dropout are applied during training to avoid artifacts when trajectories terminate early. We use 16 two NVIDIA A100 GPUsfor inference, and the whole evaluation time is around 8h.

To reduce the risk from generated artifacts, COW always retains the original image as the reliable first-frame spatial anchor, and uses generated frames only as auxiliary future-state evidence. Thus, COW is generator-agnostic in design, while its gain magnitude depends on video quality and temporal coherence. 

% \subsection{Evaluation Details}
% All the images and video frames are resized to a resolution of $512\times 512$ to ensure inference efficiency. Three frames are  

\section{Additional Analysis of \method}
\label{sec:app-iwm-exp}
To further understand the behavior and limitations of our framework, we conduct an extended analysis of \method beyond the main experiments. This section provides a deeper examination of how the model behaves under diverse spatial configurations, including both successful and problematic scenarios. We first present qualitative visualizations, and then analyze failure cases in Collision and Trajectory cases.

\subsection{Visualization}
\label{sec:ap-vis}
We provide additional qualitative results in this section to further illustrate the behavior of \method across a variety of settings. As shown in the examples in Fig.~\ref{fig:ap-visualization}, \method generates plausible and coherent video rollouts that faithfully reflect the parsed trajectories. Even in visually cluttered or geometrically complex scenes, the method demonstrates strong instance-level controllability, accurately preserving object identity, motion direction, and interaction patterns throughout the simulated sequence. Moreover, the generated rollouts exhibit high physical plausibility, as shown in the billiards example: different balls move along feasible paths, maintain realistic collision dynamics, and satisfy basic physical constraints of collision responses.

These improvements underscore the central role of grounded visual evidence during test-time reasoning. When models are provided with short simulated rollouts that explicitly encode object geometry, spatial layout, and anticipated interactions, their reasoning becomes more consistent and less reliant on implicit world assumptions. The results also demonstrate that richer visual context strengthens the benefits of test-time scaling strategies, enabling models to reason more reliably under ambiguity.

\subsection{Additional Discussion about COW}
Motivated by the failure analysis in Sec.~\ref{experiment:failures}, where MLLMs produce linguistically fluent but visually ungrounded predictions by relying on textual CoT for internal simulation, COW externalizes this process into explicit visual form to test a core hypothesis: \textit{whether grounding reasoning in explicit visual dynamics improves causal spatial inference}. Critically, COW is a \textbf{diagnostic} tool designed to probe MLLM failure modes, not a standalone solution.

CausalSpatial demands two complementary but distinct capabilities: (1) \textbf{causal simulation}: predicting future dynamics under hypothetical interventions, and (2) \textbf{spatially grounded reasoning}: interpreting fine-grained 3D geometry for final decisions. COW targets the former by supplying explicit future-state information, while the latter remains an independent bottleneck regardless of simulation quality~\cite{wan2025wan,huang2025spatial,spatial457,chen2024spatialvlm,3dsrbench}. By decoupling these two components, varying simulation quality from generated to ground-truth frames, COW directly quantifies each component's contribution to overall performance, providing principled diagnostic insight into where current systems fail. Operating entirely at \textbf{test time without additional training}, COW achieves consistent zero-shot improvements with only two to three simulated frames, though its effectiveness remains bounded by current video generation quality. This limitation also accounts for the slight regression on Occlusion, where short clips cannot capture long-horizon dynamics. As both video generation and spatial interpretation capabilities improve, COW-style frameworks are expected to yield greater gains.

\section{Data Examples}
\label{sec:examples}
\begin{figure}[h]
    \centering
    \includegraphics[width=\linewidth]{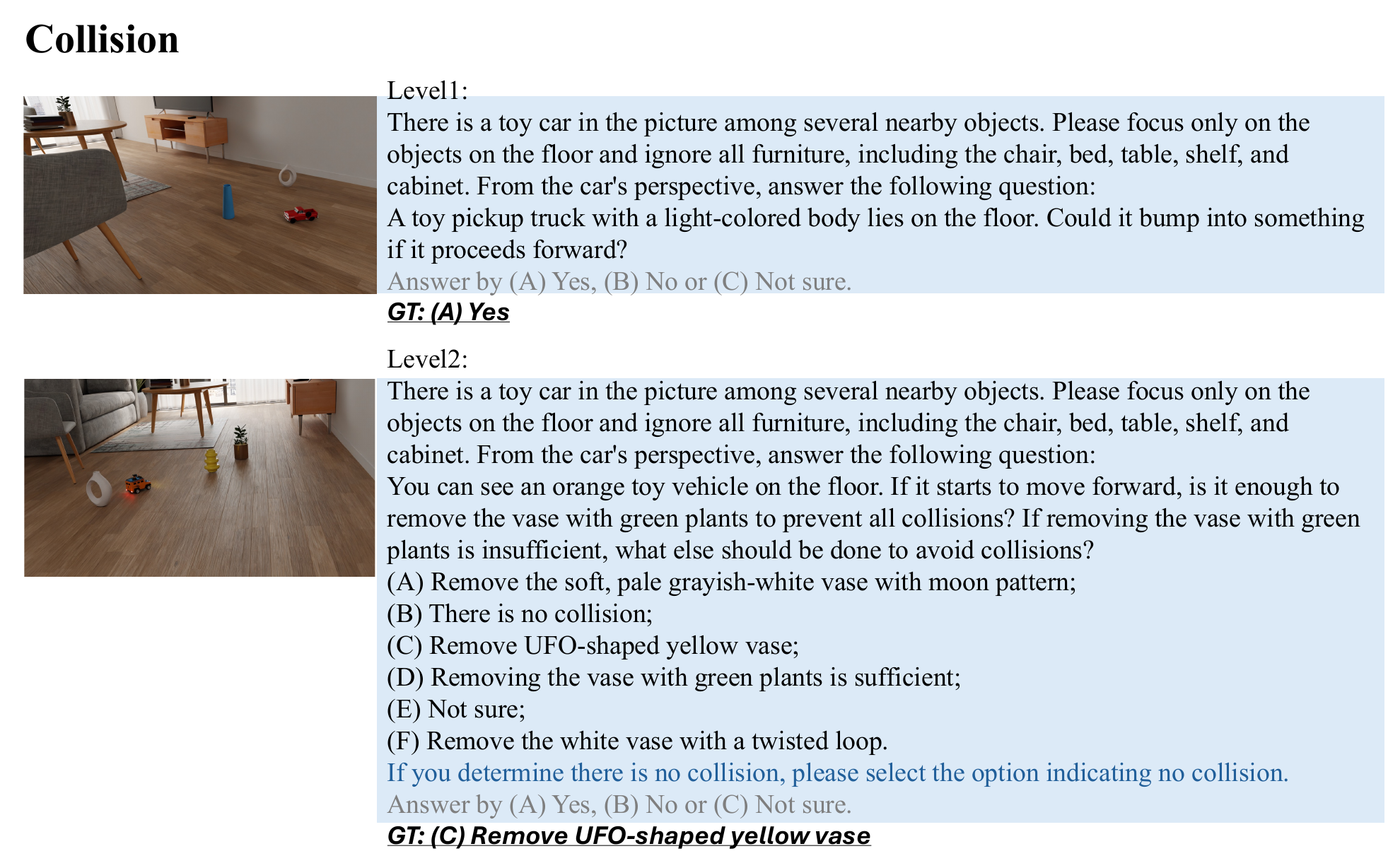}
    \caption{Examples of counterfactual collision reasoning in \bench.}
    \label{fig:ap-collision}
\end{figure}
\begin{figure}[h]
    \centering
    \includegraphics[width=\linewidth]{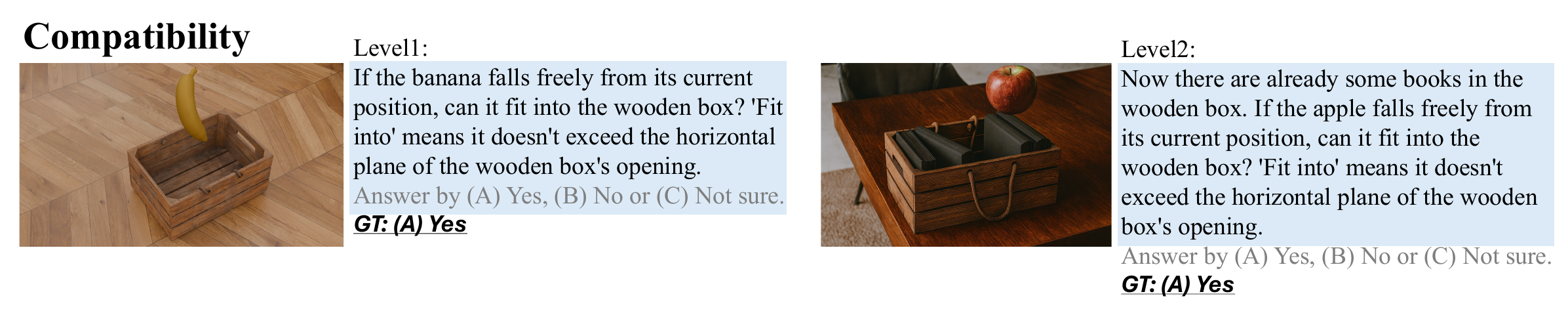}
    \caption{Examples of hypothetical compatibility reasoning in \bench.}
    \label{fig:ap-compatibility}
\end{figure}
\begin{figure}[h]
    \centering
    \includegraphics[width=\linewidth]{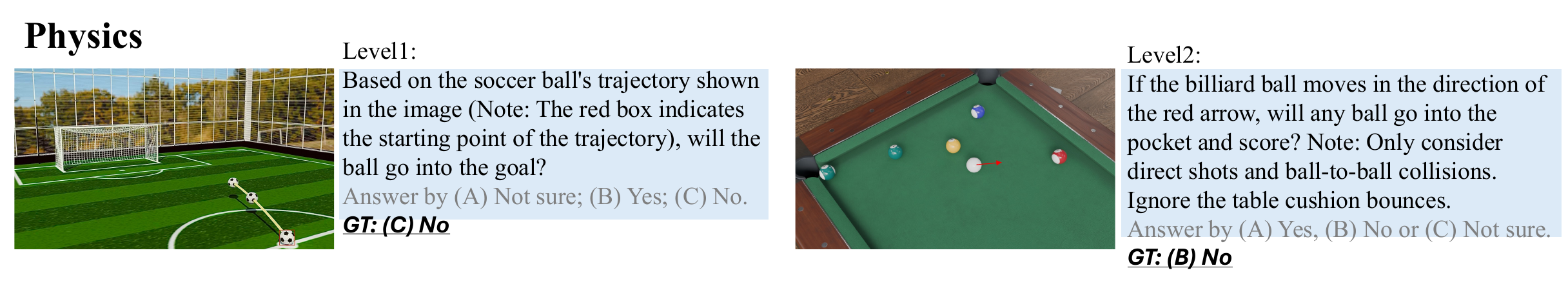}
    \caption{Examples of physics-grounded spatial anticipation task in \bench.}
    \label{fig:ap-physics}
\end{figure}
\begin{figure}[h]
    \centering
    \includegraphics[width=\linewidth]{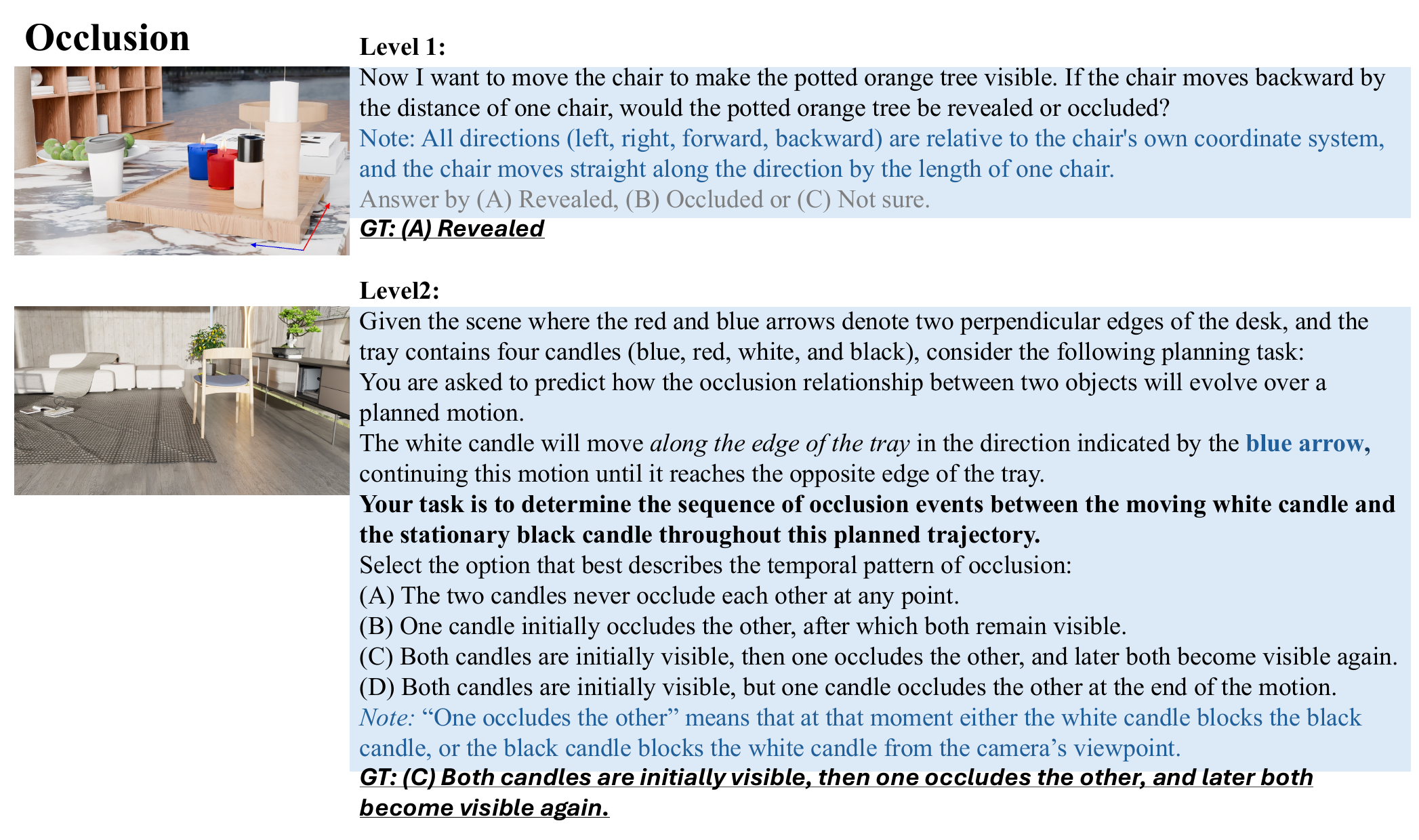}
    \caption{Examples of planning tasks with occlusion reasoning in \bench.}
    \label{fig:ap-occlusion}
\end{figure}

To illustrate the diversity and structure of our benchmark, we provide representative examples from each of the four core causal spatial reasoning tasks: collision (in Fig.~\ref{fig:ap-collision}), compatibility (in Fig.~\ref{fig:ap-compatibility}), occlusion (in Fig.~\ref{fig:ap-occlusion}), and physics-based interaction (in Fig.~\ref{fig:ap-physics}). Each example contains (i) the rendered scene, (ii) the queried causal spatial question, and (iii) the corresponding ground-truth outcome used for evaluation. Examples from both \textbf{Level1 (one-to-one)} and \textbf{Level2 (one-to-multiple)} are provided. Together, these examples demonstrate the range of object arrangements, geometric configurations, and causal dependencies present in \bench.

\subsection{Question generation}
\begin{figure}[h]
    \centering
    \includegraphics[width=\linewidth]{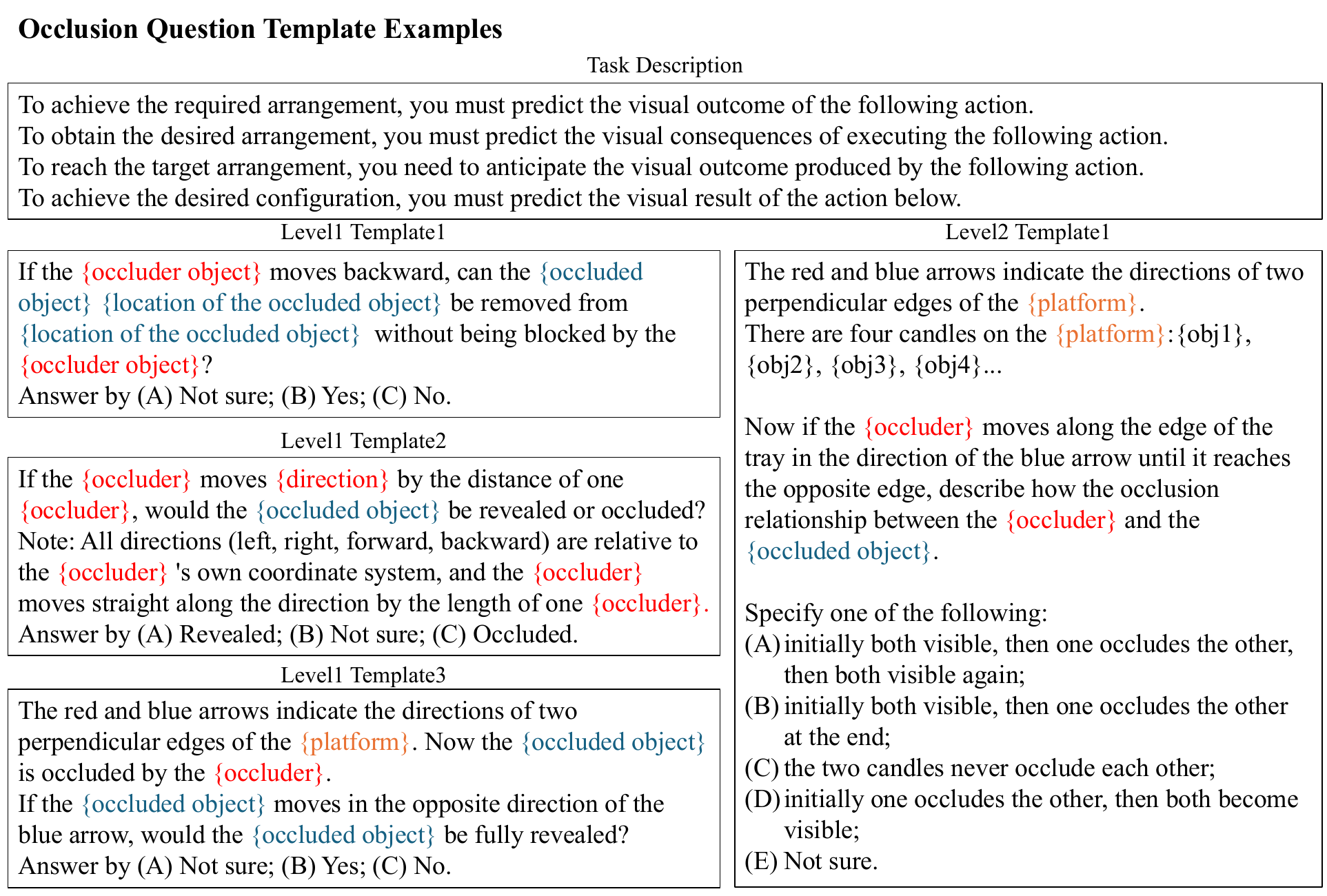}
    \caption{Examples of occlusion question templates in \bench. Each question is composed by a sampled task description and a sampled question either from L1 or L2. L1 has three different templates. }
    \label{fig:ap-occ-question}
\end{figure}

% %%%%%%%%%%%%%%%%%%%%%%%%%%%%%%%%%%%%%%%%%%%%%%%%%%%%%%%%%%%%
% \clearpage
% \newpage
% \input{checklist.tex}

\end{document}